\documentclass[times, review, 10pt]{elsarticle}
\usepackage{amssymb}
\usepackage{amsmath}
\usepackage{booktabs}
\usepackage{caption}
\usepackage{xcolor}    % soul이 색을 쓰려면 필요할 수 있음

\newcommand{\hl}[1]{#1}
\usepackage{etoolbox} % <-- patchcmd를 위한 패키지
\usepackage{adjustbox} % 크기 조절용
\usepackage{amsmath}
\usepackage{enumitem}
\usepackage{amsmath, amssymb}
\usepackage{graphicx}
\usepackage[english]{babel}
\usepackage{multirow}
\usepackage{booktabs}
\usepackage{array}

\usepackage{algorithm}        % 알고리즘 플로팅
\usepackage[noend]{algpseudocode} % 알고리즘 의사코드
\algrenewtext{EndFor}{\textbf{end for}}
\algrenewtext{EndRepeat}{\textbf{end repeat}} % Repeat용 종료 텍스트
\usepackage{bm}               % 볼드 수식 기호
\usepackage{setspace}

\newcolumntype{Y}[1]{>{\centering\arraybackslash}m{#1}}

\newcommand{\msd}[2]{\ensuremath{#1\ \pm\ {\fontsize{7pt}{8.5pt}\selectfont #2}}}
\newcommand{\msdb}[2]{\ensuremath{\mathbf{#1}\ \mathbf{\pm}\ {\fontsize{7pt}{8.5pt}\selectfont \mathbf{#2}}}}
\newenvironment{revblue}{\begingroup}{\endgroup}
\newlength{\reqindent}
\newcommand{\requireitem}{%
  \Statex\hspace*{-1.2em}$\cdot$\;}

\newcommand{\requirecont}{%
  \Statex\hspace*{-0.5em}}

\journal{Pattern Recognition}
\begin{document}
\begin{frontmatter}

\title{Compositional Meta-Learning for Mitigating Task Heterogeneity in Physics-Informed Neural Networks
}

\author[label1]{Beomchul Park}
\ead{sea_breeze88@korea.ac.kr}
\author[label1]{Minsu Koh}
\ead{minsukoh@korea.ac.kr}
\author[label2]{Heejo Kong}
\ead{hj_kong@korea.ac.kr}
\author[label1]{Seong-Whan~Lee\corref{cor1}}
\ead{sw.lee@korea.ac.kr}

\affiliation[label1]{organization={Department of Artificial Intelligence, Korea University},
            addressline={Anam-dong, Seongbuk-gu}, 
            city={Seoul},
            postcode={02841},
            country={Republic of Korea}}
            
\affiliation[label2]{organization={Department of Brain and Cognitive Engineering, Korea University},
            addressline={Anam-dong, Seongbuk-gu}, 
            city={Seoul},
            postcode={02841},
            country={Republic of Korea}}
            
\cortext[cor1]{Corresponding author}

%%%%%

%% Abstract
\begin{abstract}
\hl{Physics-informed neural networks (PINNs) approximate solutions of partial differential equations (PDEs) by embedding physical laws into the loss function. In parameterized PDE families, variations in coefficients or boundary/initial conditions define distinct tasks. This makes training individual PINNs for each task computationally prohibitive, while cross-task transfer can be sensitive to task heterogeneity. While meta-learning can reduce retraining cost, existing methods often rely on a single global initialization and may suffer from negative transfer, particularly under feature-scarce coordinate inputs and limited training-task availability. We propose the Learning-Affinity Adaptive Modular Physics-Informed Neural Network (LAM-PINN), a compositional framework that leverages task-specific learning dynamics. LAM-PINN combines PDE parameters with learning-affinity metrics from brief transfer sessions to construct a task representation and cluster tasks even with coordinate-only inputs. It decomposes the model into cluster-specialized subnetworks and a shared meta network, and learns routing weights to selectively reuse modules instead of relying on a single global initialization. Across three PDE benchmarks, LAM-PINN achieves an average 19.7-fold reduction in mean squared error (MSE) on unseen tasks using only 10\% of the training iterations required by conventional PINNs. These results indicate its effectiveness for generalization to unseen configurations within bounded design spaces of parameterized PDE families in resource-constrained engineering settings.}

\end{abstract}

% %% Keywords

\end{frontmatter}



\section{Introduction}

Partial differential equations (PDEs) are widely used to model physical systems. Physics-informed neural networks (PINNs) approximate PDE solutions by minimizing a loss that combines data mismatch and PDE-residual penalties evaluated at collocation points via automatic differentiation \cite{raissi2019physics}. This formulation enables mesh-free solution approximation from sparse measurements and has been applied to engineering problems such as fluid dynamics and structural analysis \cite{raissi2019physics,tu2022physics}, as well as to image processing tasks \cite{xie2024node, mao2025pid}.
\begin{revblue}

In practice, engineering workflows often require solving many instances of a parameterized PDE. Even within a single PDE family, variations in coefficients, material parameters, or boundary/initial conditions define distinct tasks. Retraining a PINN for each task can be prohibitive for design exploration and optimization. Meta-learning and transfer learning aim to amortize this cost by learning parameters that adapt rapidly to new configurations \cite{lu2023survey}. Optimization-based methods such as MAML \cite{finn2017model} and its PINN adaptations \cite{liu2022novel} are representative baselines. However, they typically transfer knowledge through a single global initialization. When the training tasks are heterogeneous, this strategy can induce negative transfer.

This issue is especially relevant in PINNs. In contrast to feature-rich few-shot benchmarks, PINN tasks use coordinate-only inputs and often provide only a limited number of training tasks. The available task set may be limited in size, yet heterogeneous due to the diverse combinations of governing conditions. Fig.~\ref{fig:fig1} illustrates this mismatch. Within the design-of-experiments (DoE) range, two unseen Helmholtz tasks share the same PDE form and differ only in parameter values, yet each baseline exhibits markedly different transfer performance across the two tasks.

\begin{figure}[t]
\centering
\includegraphics[width=0.65\linewidth]{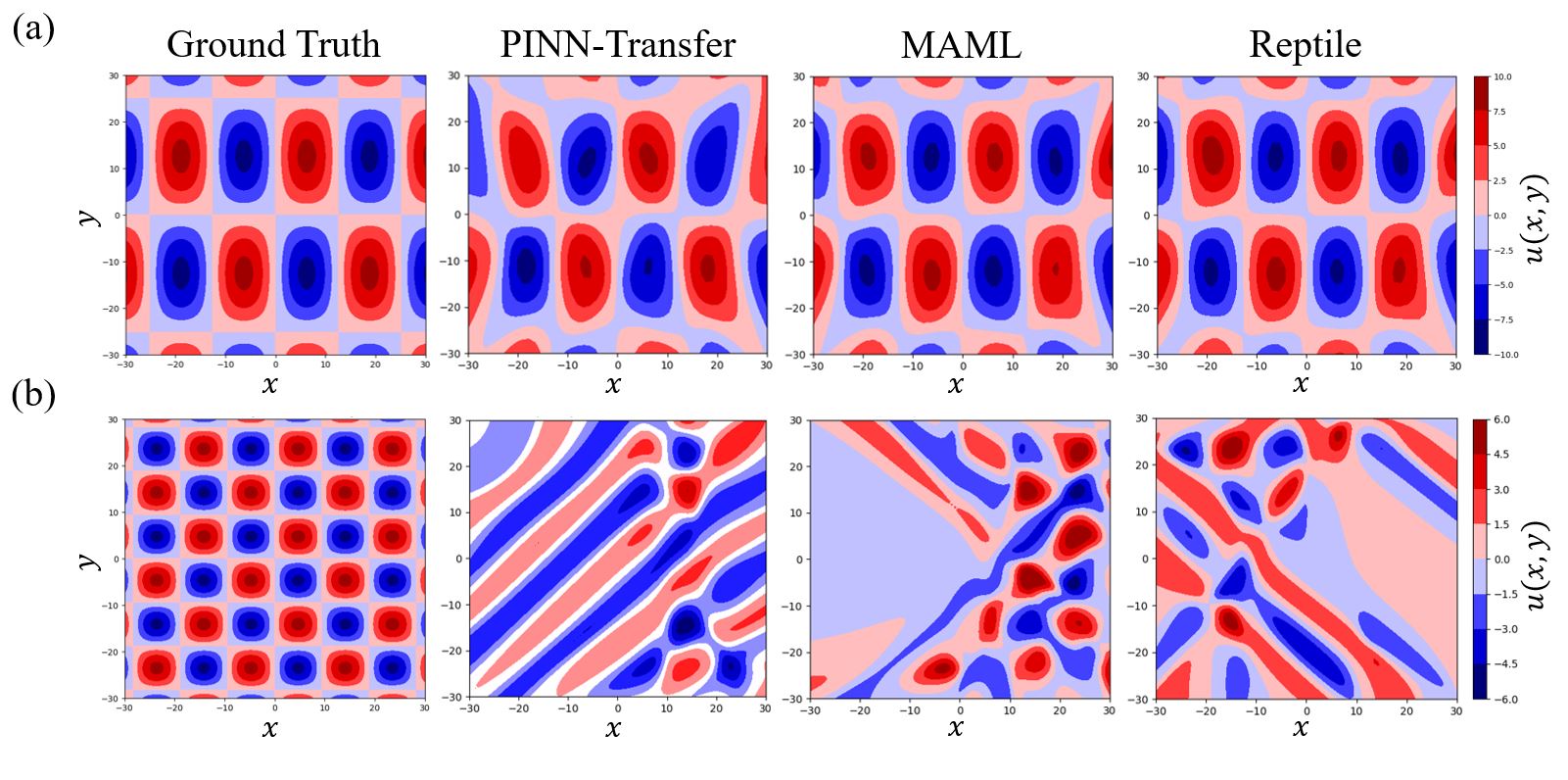}
\caption{
Transfer performance differences of pre-trained models on distinct unseen tasks highlight the impact of task heterogeneity. Subfigures (a) and (b) show the results of each baseline model on two unseen tasks governed by the Helmholtz equation, differing only in the equation parameters.
Despite identical architectures and training settings, the results for task (b) consistently show lower transfer performance than those for task (a).
}
\label{fig:fig1}
\end{figure}

Task-aware meta-learning mitigates heterogeneity by explicitly modeling task identity and learning cluster-specialized components via task clustering \cite{lin2021task,peng2023clustered}. However, directly applying these ideas to PINNs remains difficult. Coordinate-only inputs provide little task-discriminative information, and descriptors based only on PDE parameters \cite{huang2022meta,cho2023hypernetwork} may not capture task-specific learning dynamics. Accordingly, an important remaining challenge is how to explicitly account for task heterogeneity in PINNs for parameterized PDE families, particularly under coordinate-only inputs and a limited training-task regime.

This study focuses on generalization to unseen parameter configurations within a bounded DoE range of the same PDE family, where a range of parameter configurations may give rise to a heterogeneous task setting. Within this scope, we examine three questions: how to represent task heterogeneity when the input does not directly reveal task identity; which parts of a PINN should be specialized or shared to reduce negative transfer while preserving transfer efficiency; and whether such selective reuse improves adaptation to unseen tasks under limited training budgets.

We address these questions with the Learning-Affinity Adaptive Modular Physics-Informed Neural Network (LAM-PINN), a compositional meta-learning framework for task-heterogeneous PINNs. LAM-PINN represents each task by combining PDE parameters with learning-affinity metrics from brief transfer sessions, capturing early-stage dynamics to enable clustering even under coordinate-only inputs. Based on layer-wise analysis, we specialize input-adjacent layers into cluster-specific modules while sharing the remaining layers as a meta-network. During adaptation, learnable routing weights compose these modules into a task-specific initialization, aiming to shift the starting point in the parameter space toward the target task for more efficient convergence (Fig.~\ref{fig:fig2}b).

\begin{figure}[t]
\centering
\includegraphics[width=0.6\linewidth]{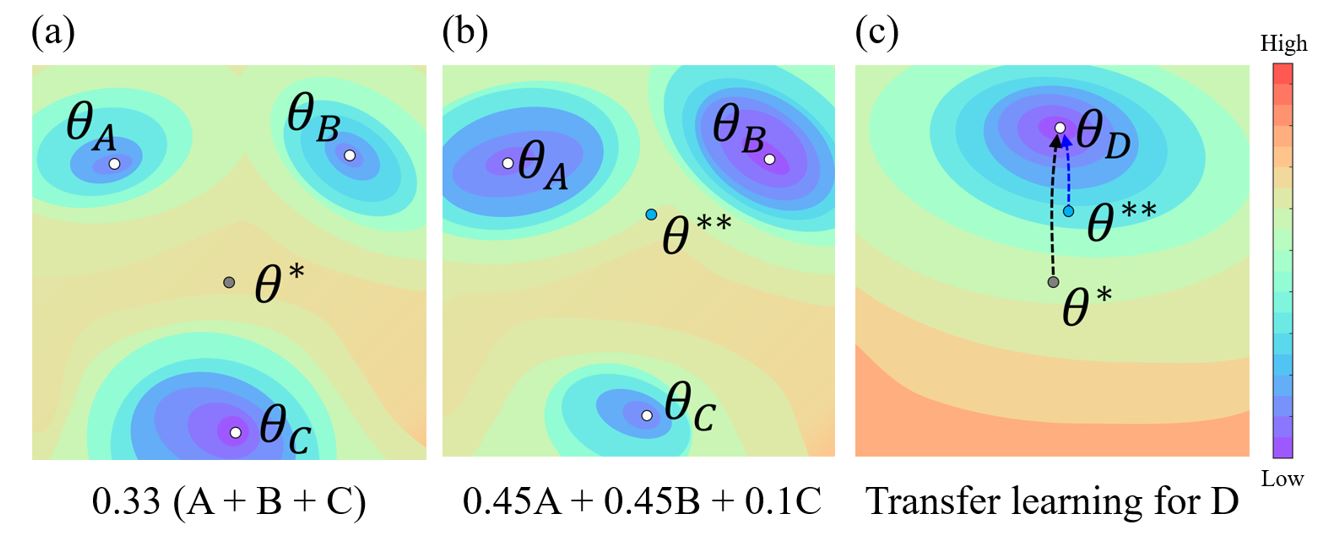}
\caption{
Conceptual diagram of the proposed method in parameter space. Boxes represent the high-dimensional parameter space of the neural network, and contours indicate loss or energy levels.
(a) Optimal weights of tasks A, B, and C, and a meta-learned initialization $\theta^*$.
(b) Meta-initialization adjusted based on task-specific contributions.
(c) Comparison of adaptation paths to an unseen task D from two different initializations.
}
\label{fig:fig2}
\end{figure}

To reflect resource-constrained engineering workflows, we form the task set using a three-factor, three-level full-factorial DoE over PDE coefficients, initial conditions (ICs), and boundary conditions (BCs), resulting in 27 training tasks \cite{heckert2002handbook}. We evaluate LAM-PINN on three representative PDE benchmarks, comparing its performance against standard optimization-based and contrastive-based meta-learning baselines, alongside recent PINN-oriented adaptations. We also discuss limitations and failure cases to clarify when selective reuse is ineffective. The primary contributions are summarized below.

\begin{enumerate}[label=\arabic*), itemsep=0pt]
\item We introduce a learning-affinity task representation for PINNs that combines PDE parameters with metrics from brief transfer sessions, yielding informative task embeddings under coordinate-only inputs with minimal additional overhead.

\item We propose a modular PINN architecture that specializes only input-adjacent layers into cluster-specific subnetworks and composes them with a shared meta network via learnable routing. With 27 DoE training tasks, it achieves an average 19.7-fold reduction in MSE on unseen tasks while using  ~10\% of the per-task training iterations required by conventional PINNs.
\end{enumerate}
\end{revblue}

\section{Background and Motivating Observations}
\subsection{Physics-informed neural networks and parameterized PDEs}

This subsection reviews the standard PINN objective for parameterized PDE families, which we use for modular training and rapid adaptation (Sec.~3.3--3.4). In our framework, a task $\tau$ corresponds to a specific PDE configuration (e.g., coefficients, ICs, and BCs). Given a network prediction $u(\cdot;\theta)$, a PINN minimizes

\begin{equation}
L(\theta) = L_{\text{data}}(\theta) + L_{\text{physics}}(\theta),
\end{equation}
where $L_{\text{data}}$ encourages the network prediction to match $N$ labeled measurements:
\begin{equation}
L_{\text{data}}(\theta) = \frac{1}{N} \sum_{i=1}^{N} \left| u(x_i, t_i; \theta) - u_i \right|^2,
\end{equation}
Here, $\{(x_i,t_i,u_i)\}_{i=1}^N$ denotes the labeled measurements, with $(x_i,t_i)$ as the input coordinates and $u_i$ as the observed value. Similarly, $L_{\text{physics}}$ enforces the governing PDE by penalizing the residual at $M$ collocation points $\{(x_j,t_j)\}_{j=1}^M$:
\begin{equation}
L_{\text{physics}}(\theta) = \frac{1}{M} \sum_{j=1}^{M} \left| \mathcal{F}\!\left(u(x_j, t_j; \theta)\right) \right|^2.
\end{equation}

The collocation points are unlabeled coordinates used to enforce the PDE constraints through the residual. 
The operator $\mathcal{F}$ denotes the PDE residual evaluated via automatic differentiation. Although Eqs.~(1)--(3) share the same form across tasks, changes in the PDE configuration (coefficients, BCs/ICs) induce heterogeneous learning dynamics, which LAM-PINN mitigates through modular specialization.

\subsection{Generalization in PINNs}

To address the limited generalization capability of PINNs, various transfer-learning approaches have been explored \cite{5288526}. Existing strategies include one-shot inference via matrix factorization \cite{desai2021one}, hybrid schemes that couple pre-trained PINNs with multi-fidelity datasets \cite{chakraborty2021transfer}, and progressive transfer procedures that account for the relative learning difficulty across PDEs \cite{krishnapriyan2021characterizing}. Myung et al.\ further improve transfer-time efficiency by pruning weights during adaptation while preserving conservation-related physical priors \cite{myung2022pac}.

Optimization-based meta-learning methods such as MAML and REPTILE have also been adapted to PINNs to learn an initialization that rapidly adapts to unseen PDE configurations with limited data \cite{liu2022novel, voon2025trapezoidal}. Recent work further accounts for configuration-dependent difficulty by introducing a difficulty-aware task sampler (DATS) \cite{toloubidokhti2024dats}. In parallel, parameter-conditioned and hypernetwork-based PINNs condition the solution network on PDE parameters. Meta-Auto-Decoder (MAD) and parameterized PINNs (P$^{2}$INN) map PDE parameters to low-dimensional latent codes that guide adaptation \cite{huang2022meta, cho2024ppinn}, while Hyper-LR-PINN uses a lightweight hypernetwork to output layer-wise diagonal coefficients for a low-rank PINN \cite{cho2023hypernetwork}. While effective, these auxiliary networks increase architectural complexity and can add optimization overhead. Recent benchmarking efforts broaden the comparison set by including neural operators \cite{yee2026meta}. Relatedly, meta-learning across different PDE problems via problem encoding and forward-pass prediction has also been studied \cite{iwata2023meta}.

\subsection{Task-aware meta-learning and heterogeneity}

Beyond global meta-initializations, task-aware meta-learning explicitly models task identity or similarity. Representative directions include probabilistic formulations that model inter-task uncertainty \cite{yoon2018bayesian}, uncertainty-aware adaptation under distribution shift \cite{neupane2021metaedl}, feature-conditioned adaptation networks \cite{requeima2019fast}, and task-dependent modulation mechanisms \cite{oreshkin2018tadam}. Recent work also emphasizes explicit task representations derived from task features or learning signals, including initialization modulation \cite{vuorio2019multimodal}, $k$-means grouping in representation space \cite{lin2021task}, and training dynamics such as learning trajectories or gradients \cite{peng2023clustered, mu2020gradients}. \hl{ In a similar vein, ConML \cite{wu2025learning} employs a task-level contrastive objective in model space, aligning representations of the same task while discriminating between different tasks.}

Several studies explicitly address heterogeneity-induced negative transfer and interference. Iwata and Kumagai infer permutation-invariant latent vectors for tasks with heterogeneous attribute spaces \cite{iwata2020meta}. Wang et al.\ mitigate negative interference using task-specific adapters trained with a bilevel meta-learning objective \cite{wang2020negative}, and further reduce confounder-driven spurious correlations by disentangling generating factors and enforcing support--query invariance \cite{wang2023hacking}.

Compositional meta-learning provides another route by composing reusable components via routing or gating \cite{bakermans2025compositional}. However, transferring these strategies to PINNs is nontrivial because the inputs are fixed low-dimensional coordinates, whereas task variation arises from operator-level physics (coefficients, ICs, BCs, and geometry), and both task and measurement budgets are often limited. This setting can amplify negative transfer under a single global meta-initialization, and DATS further highlights the importance of configuration-dependent difficulty in meta-PINNs \cite{toloubidokhti2024dats}. These observations motivate task representations derived from learning behavior rather than input statistics, together with selective reuse mechanisms tailored to PINN settings.

\begin{figure}[t!]
\centering
\includegraphics[width=1\linewidth]{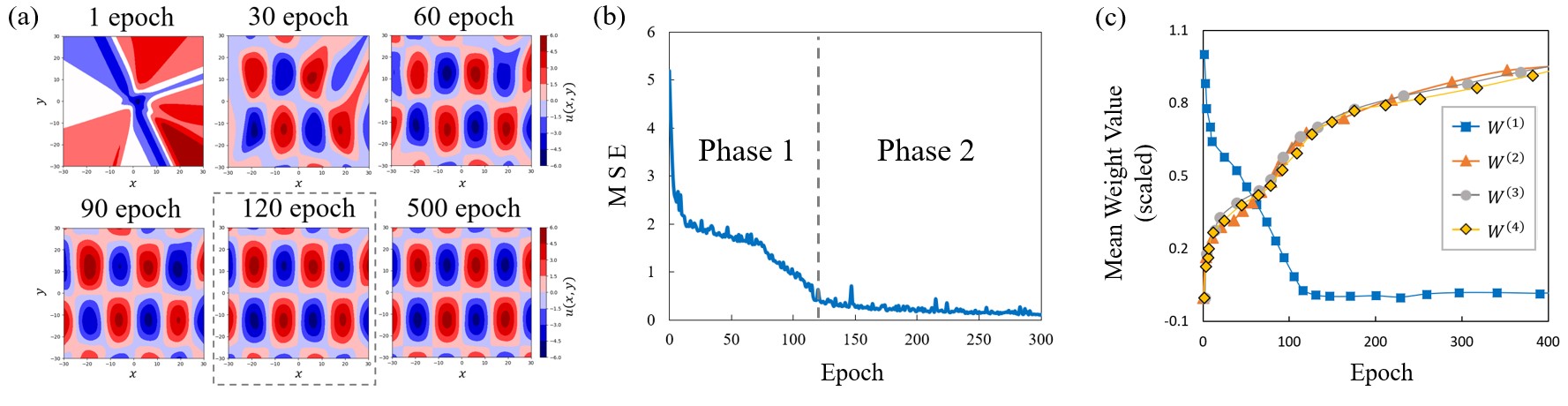}
\caption{
 Transfer learning analysis of PINNs. (a) Visualization of predicted solutions during transfer learning on the Helmholtz equation. (b) MSE curves over training iterations. (c) Evolution of layer-wise average weight magnitudes throughout training.
}
\label{fig:fig3}
\end{figure}

\subsection{Layer-wise learning dynamics and motivation for modularization}

This subsection identifies which layers drive fast transfer in PINNs and motivates our modular split. While early layers of conventional MLPs often learn generic features and deeper layers capture task-specific abstractions \cite{yosinski2014transferable,montufar2014number}, PINNs use coordinate-based neural fields \cite{cao2025nerf,czerkawski2024neural} to map low-dimensional coordinates to continuous fields. Under PDE-residual constraints, the transfer-critical layers are less obvious.

We study Helmholtz transfer using a PINN with one input layer, three hidden layers, and one output layer. Transfer exhibits a two-phase behavior (Fig.~\ref{fig:fig3}): the coarse spatial pattern emerges quickly (Fig.~\ref{fig:fig3}(a)), followed by slower refinement, reflected by an early steep MSE drop and a later plateauing regime (Fig.~\ref{fig:fig3}(b)). To localize this behavior in the network, we track layer-group weight magnitudes $W^{(1)}$--$W^{(4)}$ (input-to-first hidden through near-output). As shown in Fig.~\ref{fig:fig3}(c), $W^{(1)}$ changes rapidly at the start of transfer and stabilizes early, whereas deeper groups keep evolving, suggesting that input-adjacent layers capture coarse structure while later layers refine details.

\begin{figure}[t!]
\centering
\includegraphics[width=1\linewidth]{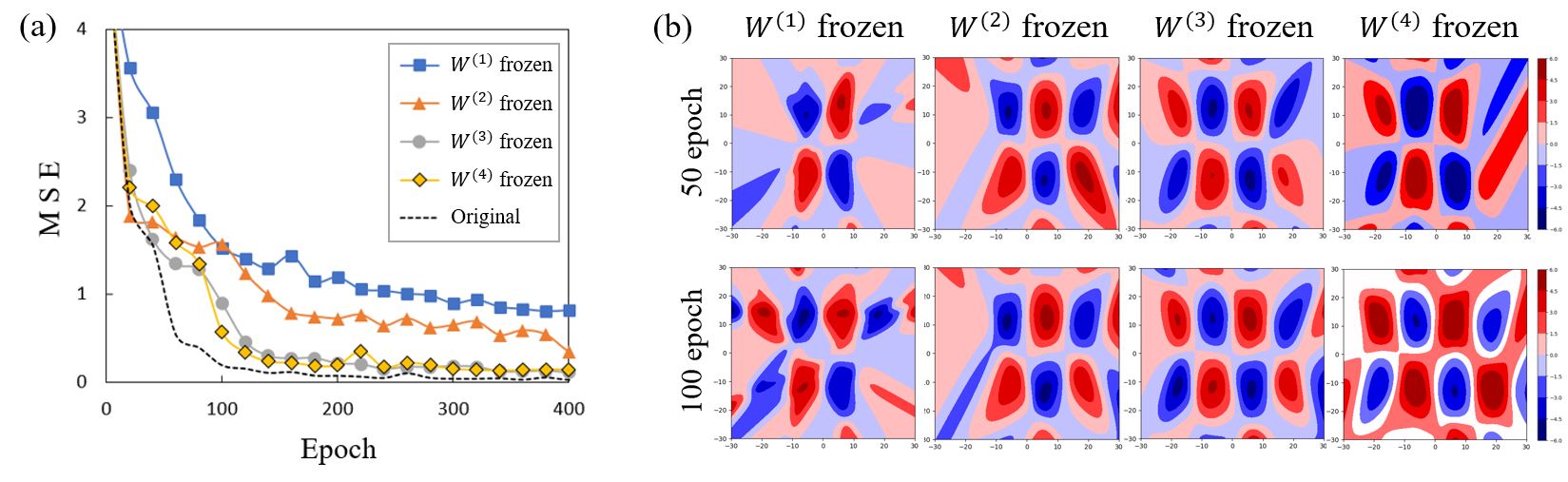}
\caption{
Learning trends and results with layer freezing. (a) MSE graph during transfer learning using the Helmholtz equation. (b) Inferred images at 50 and 100 epochs.
}
\label{fig:fig4}
\end{figure}

We validate this with layer-freezing. Freezing early layers, especially $W^{(1)}$, markedly slows error reduction (Fig.~\ref{fig:fig4}(a)) and degrades reconstructed fields at 50 and 100 epochs (Fig.~\ref{fig:fig4}(b)), whereas freezing deeper layers has a smaller effect. Therefore, LAM-PINN modularizes only the early, input-adjacent layers into cluster-specific subnetworks and shares the remaining layers as a meta network for refinement, defining the modular boundary used in Sec.~3.

\section{The Proposed Method}

\begin{figure}[t!] \centering \includegraphics[width=0.8\linewidth]{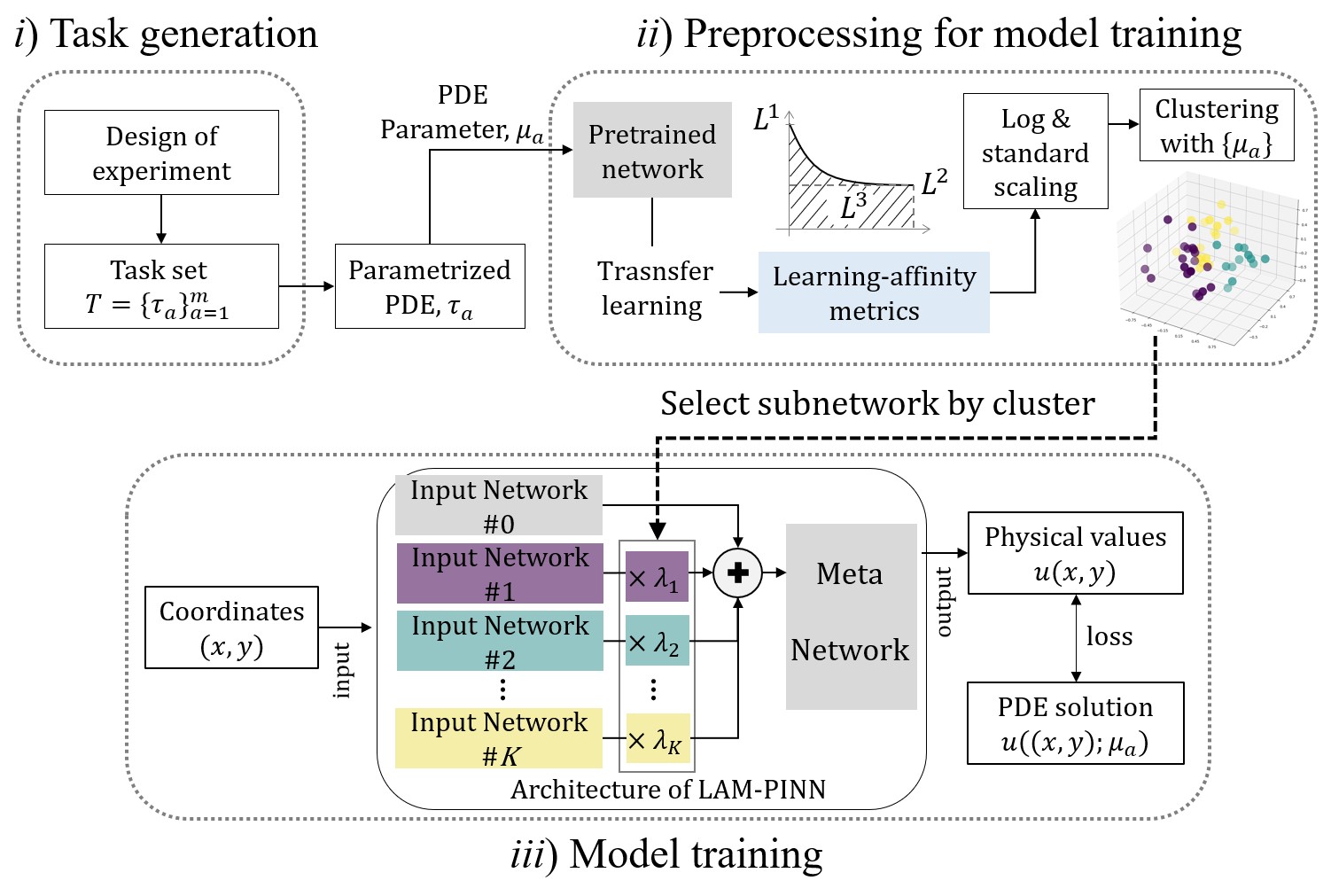} \caption{ Schematic of LAM-PINN training process. \textit{i}) Task generation using DoE; \textit{ii}) Preprocessing to structure and cluster tasks based on learning-affinity metrics; and \textit{iii}) Model training, with clustered tasks learned separately. } \label{fig:fig5} \end{figure}

As illustrated in Fig.~\ref{fig:fig5}, the proposed LAM-PINN framework consists of four phases: \textit{i}) task generation using DoE, \textit{ii}) preprocessing for learning-affinity-based clustering, \textit{iii}) modular training of subnetworks, and \textit{iv}) adaptive transfer to unseen tasks.

\subsection{Task generation with DoE}
This subsection constructs the training task set $T$ by generating diverse PDE configurations via a full-factorial DoE. DoE is a systematic approach for planning, conducting, and analyzing controlled tests to evaluate the effects of various factors on outcomes \cite{ganesan1995statistical}. By systematically varying experimental factors, DoE covers the task space while reducing the number of required experiments. We use a full factorial design (FFD) to enumerate all combinations of factor levels and capture their interactions. We represent each task by a task configuration vector $\boldsymbol{\mu}=\{\mu^p\}_{p=1}^{P}$, whose elements specify the task-defining quantities of the target equation---PDE parameters, ICs, and BCs---with $p$ indexing the parameter type.  For each $\mu^p$, we specify a range $\mu^p \in [\mu^p_{\min}, \mu^p_{\max}]$, and generate tasks by combining these settings. The resulting task set is $T = \{\tau_a\}_{a=1}^m$, where $m$ is the number of tasks.

\subsection{Preprocessing for model training}

We construct a task representation for clustering by combining PDE parameters with loss-based learning-dynamics signals. Because PINNs utilize feature-scarce coordinate inputs, parameter-only descriptors often fail to predict transfer performance across heterogeneous tasks. To address this, we augment PDE parameters with loss-dynamics signals extracted from a brief transfer session, using less than 5\% of typical convergence of a pre-trained reference PINN. This choice is motivated by the rapid early-phase adaptation in Fig.~\ref{fig:fig3} (coarse-to-fine learning) and is consistent with the spectral bias of neural networks \cite{rahaman2019spectral}, which suggests that coordinate-based MLPs tend to learn low-frequency, global structures before high-frequency details. Under this view, tasks within a fixed PDE family that share similar dominant solution structures are more likely to exhibit similar early-stage loss dynamics, making these signals a practical, empirically grounded proxy for task similarity. \hl{In practice, this short budget is sufficient to capture stable affinity signals for clustering; we further analyze the representation and clustering behavior in Sec.~4.3 and assess robustness with respect to the transfer-session budget and clustering stability in Sec.~4.4. More concretely, within each PDE family, we transfer the same pre-trained reference PINN to every DoE-generated task $\tau_a \in T$ for the short budget and record three scalar total-loss statistics: the initial loss $L_a^1$ (epoch 0), the final loss $L_a^2$ (end of the short session), and the epoch-averaged loss $L_a^3$ over that session. These three values are concatenated, in this fixed order, with the task-configuration vector $\boldsymbol{\mu}_a$ containing the PDE coefficients and IC/BC-related DoE variables to form the unified task embedding $f_a$:
\begin{equation}
f_a = \text{normalize} \left( \log \left( \mathbf{1} + \mathbf{Z}_a \right) \right),\quad
\mathbf{Z}_a = [\mu_a^1, \dots, \mu_a^P, L_a^1, L_a^2, L_a^3].
\end{equation}
The element-wise $\log(\mathbf{1}+\mathbf{Z}_a)$ transformation mitigates scale disparity and reduces outlier effects, and the subsequent Z-score normalization is applied feature-wise over the DoE task set before $k$-means clustering.} Finally, we group the tasks $\{f_a\}_{a=1}^{m}$ using $k$-means clustering \cite{franti2019much} by minimizing the within-cluster variance:
\begin{equation}
\underset{C}{\arg\min} \sum_{j=1}^{K} \sum_{f_a \in C_j} \left\| f_a - \phi_j \right\|^2,
\end{equation}
where $\phi_j$ denotes the centroid of cluster $C_j$. In this embedding space, Euclidean proximity can be used as a proxy for inter-task learning affinity. The resulting clustered task sets $\{T_j\}_{j=1}^{K}$ provide the structural basis for training the specialized subnetworks detailed in Sec.~3.3.

% %────────────────────────────────────────────────────────────────────────────────────────────
% %  의사코드
% %───────────────────────

\begin{algorithm}[t]
\caption{The Proposed Framework}
\centering
\resizebox{0.85\linewidth}{!}{%
\begin{minipage}{\linewidth}
\small
\begin{algorithmic}[1]

\Require
\requireitem Task set from DoE ${T}=\{\tau_a\}_{a=1}^{m}$, $m$ = number of tasks
\requireitem $\theta_{\tau_1}$: pre-trained model with equation parameters from $\tau_1$
\requireitem Split $\theta_{\tau_1}$ into 
            $\theta^{0}_{\text{IN}} = \{\theta^{1}_{\tau_1}, \dots, \theta^{l^\ast}_{\tau_1}\}$,
            $\theta_{\text{MN}} = \{\theta^{l^\ast+1}_{\tau_1}, \dots, \theta^{l}_{\tau_1}\}$
\requirecont where $l^\ast=$ IN depth, $l=$ total model depth
\requireitem $K$: number of clusters based on learning-affinity metrics
\requirecont and $\mu$ values
\requireitem Task set divided into clusters ${T}_j,\; j=1,\dots,K$
\requireitem $N_1,N_2$: number of learning iterations

\State Randomly initialize all $\theta^{j}_{\text{IN}}\;(j=1,\dots,K)$
\Repeat \ \ for $n$ epochs :
  \For{$j = 1,2,\dots,K$} \        \ \ \ \#\ Phase 1
    \State Sample task $\tau_i \sim {T}_j$
    \For{$\text{iteration} = 1,2,\dots,N_1$}
       \State $\lambda_j \ = 1$, others $\lambda \ = 0.1$
       \State Update $\theta \gets \theta - \alpha\nabla_{\theta} L_{\tau_i}(\theta)$
    \EndFor
  \EndFor
  \State Freeze all $\theta^{0}_{\text{IN}}$ \& $\theta^{j}_{\text{IN}}$ \ \ \ \ \ \ \ \ \#\ Phase 2
  \For{$\text{iteration} = 1,2,\dots,N_2$}
    \For{$j = 1,2,\dots,K$}
       \State Sample task $\tau_i \sim {T}_j$
       \State $\lambda_j \ = 1$, others $\lambda \ = 0.1$
       \State Compute $L_{\tau_i\sim{T}_j}(\theta)$
    \EndFor
    \State Update $\theta \gets \theta - \alpha
            \nabla_{\theta}\displaystyle\sum_{j=1}^{K}
            L_{\tau_i\sim{T}_j}(\theta)$
  \EndFor
  \State Unfreeze all $\theta^{0}_{\text{IN}}$ \& $\theta^{j}_{\text{IN}}$
\Until{all epochs are completed}

\end{algorithmic}
\end{minipage}}
\end{algorithm}
% %────────────────────────────────────────────────────────────────────────────────────────────

\subsection{Model training}
This section describes the LAM-PINN architecture and training process. As shown in Fig.~\ref{fig:fig5}, the framework comprises one meta network (MN) and $(1+K)$ input networks (INs), where $K$ is the number of clusters. To leverage the transfer learning properties of PINNs, the early-layer weights near the input, denoted as ${\theta}_{\text{IN}}$, are separated from the pre-trained model (${\theta}_{\text{Total}}$) used in the preprocessing phase. These weights are parallelized into multiple subnetworks of equal size, ${\theta}_{\text{IN}}^0$ and ${\theta}_{\text{IN}}^j$ $(j=1,\dots,K)$, to enable learning adapted to specific task properties. The weights of the later layers near the output, represented as ${\theta}_{\text{MN}}$, combine the output vectors from the INs to produce the final predictions. Here, ${\theta}_{\text{MN}}$ denotes the remaining network weights not included in ${\theta}_{\text{IN}}$ (i.e., ${\theta}_{\text{MN}}={\theta}_{\text{Total}}-{\theta}_{\text{IN}}$).

The ${\theta}_{\text{IN}}^0$, derived from the pre-trained model, serves as a conservative initialization preserving fundamental physical knowledge inherent to the governing PDE. In contrast, each ${\theta}_{\text{IN}}^j$ is assigned to a specific cluster and optimized to capture the distinct learning properties of its corresponding task set. During training, the output vectors of the INs are combined through routing weights $\boldsymbol{\lambda} = (\lambda_1, \dots, \lambda_K)$ constrained to $(0, 1]$. To ensure all INs contribute to joint learning, $\lambda_j$ is never set to zero. The main IN receives a larger $\lambda_j$ (e.g., 1), while smaller values (e.g., 0.1) are assigned to others to preserve their learned knowledge. Consequently, only ${\theta}_{\text{IN}}^j$ and ${\theta}_{\text{MN}}$ are updated, whereas $\boldsymbol{\lambda}$ acts as a fixed set of scaling factors. Through this scheme, the INs are trained in a cluster-wise manner to learn task-specific features. The INs are trained for task-dependent performance, whereas the MN, inspired by MAML, is optimized for generalization to achieve task-independent precision performance. The model architecture can be expressed as follows:

\begin{align}
    {h}_{\text{IN}} &= g(x, y; {\theta}_{\text{IN}}^0) + \sum_{j=1}^K \lambda_j \cdot g(x, y; {\theta}_{\text{IN}}^j), \\
    u &= g({h}_{\text{IN}}; {\theta}_{\text{MN}}).
\end{align}

\indent Here, ${h}_{\text{IN}}$ is the sum of the output vectors from INs, and $u$ represents the predicted physical values. The function $g(\cdot\ ;\ {\theta})$ denotes the forward operator of a neural network parameterized by ${\theta}$, mapping input features to output vectors. The learning process proceeds in two phases, as detailed below.
\begin{enumerate}[label=\arabic*), itemsep=0pt]
    \item Clustered task-wise training: In this phase, each IN is trained individually on clustered tasks, with ${\theta}_{\text{IN}}^0$ fixed as the baseline. For training, tasks $\tau_i$ are sequentially sampled from each cluster $T_j$ $(j=1,\dots,K)$ to minimize the loss $L_{\tau_i}$. Scaling factors $\lambda$ are applied to modulate the contribution of each IN based on task relevance. This phase lays the foundation for efficient modular adaptation during transfer learning (detailed in Phase 1, Algorithm 1). 
    \item Meta-training: After IN specialization, the MN is trained to generalize across all clusters, with IN weights frozen to preserve cluster-specific knowledge. During this phase, tasks are resampled from each cluster, and the aggregated losses are used to update ${\theta}_{\text{MN}}$, enabling robust generalization to unseen tasks. Once training is complete, the IN weights are unfrozen (detailed in Phase 2, Algorithm 1). 
\end{enumerate}

The two training phases are repeated until all tasks in each cluster are learned, enabling the LAM-PINN to efficiently perform transfer learning by leveraging cluster-specific specialization and global generalization.

\subsection{Transfer learning with LAM-PINN}

The trained LAM-PINN framework employs $K$ independently specialized INs, each tailored to a specific task cluster, for effective transfer learning. For clarity, we denote by $\Theta$ the full set of trainable parameters in this stage, that is, $\Theta = \{\theta_{\text{IN}}, \theta_{\text{MN}}, \boldsymbol{\lambda}\}$, where $\theta_{\text{IN}}$ collects all IN parameters and $\boldsymbol{\lambda} = (\lambda_1,\dots,\lambda_K)$ are the routing weights appearing in Eq.~(6). In the transfer phase for an unseen task, we minimize the same PINN loss $L(\Theta)$ defined in Eqs.~(1)–(3) with respect to all components of $\Theta$.

Unlike the training phase, where the routing weights are kept fixed to heuristic constants, we reinitialize them to a neutral and symmetric value at the beginning of transfer, setting $\lambda_j = 0.5$ for $j = 1,\dots,K$ so that all subnetworks start from an equal contribution rather than inheriting the training-time scales. During subsequent adaptation, the routing weights are updated together with the other network parameters by gradient-based optimization. After each gradient step, we clip $\lambda_j$ into $[0, 1]$ to stabilize optimization.
\begin{align}
  \lambda_j \leftarrow \lambda_j - \alpha\,\frac{\partial L(\Theta)}{\partial \lambda_j}, 
  \qquad j = 1,\dots,K,
\end{align}
where $\alpha$ denotes the learning rate. This method uses pre-trained weights, which have been individually trained based on task properties, to allow the model to combine multiple specialized branches according to task-specific routing weights. %The proposed model overcomes the limitations of small datasets and enables more proactive transfer learning than conventional meta-learning methods.

\section{Experiments}

\subsection{Experimental design}
Higher-level designs, though more expressive, are often infeasible in practice due to prohibitive experimental costs, making a 3-level design a practical alternative \cite{heckert2002handbook}. Instead of a single design variable, we selectively combined three variables---equation parameters, ICs, and BCs---across three levels (high, medium, and low) to reflect diverse physical conditions. Accordingly, a 3-factor, 3-level factorial design was adopted, considering high experimental cost and data scarcity in real-world settings, resulting in only 27 training cases. This factorial design achieves a practical balance between modeling richness and data efficiency, and it mirrors practical environments that require multiple predictions or optimization within specific parameter ranges.

\hl{Using this DoE task set, the key design choices of LAM-PINN were fixed through small-scale pilot runs. The short transfer-session budget used to construct task embeddings was chosen from a low-budget regime (\(<5\%\) of the iterations typically required for a conventional PINN to converge), where clustering remained stable (Table~\ref{tab:helmholtz_combined}). The number of clusters was selected from \(K \in \{2,\dots,6\}\) by jointly considering cluster separation and seed-level stability. Learning rates were explored between \(1\times10^{-3}\) and \(5\times10^{-3}\), and the benchmark-specific phase-wise training budgets were then determined through pilot sweeps under the selected clustering, retaining the smallest settings that yielded stable convergence and consistent downstream transfer behavior. Exact benchmark-specific computational budgets, parameter counts, collocation-point settings, final architectures, and optimization schedules are summarized in Appendix~A, and the clustering protocol is summarized in Appendix~B. All experiments were conducted on a workstation equipped with NVIDIA GeForce RTX 3090 GPUs, an AMD Ryzen 7 3700X CPU, and 78 GiB RAM.} 

\hl{Across the experiments, we use three evaluation modes. First, fixed 10-task benchmark comparisons assess average transfer performance over a common unseen-task set shared across methods. This benchmark reflects engineering optimization settings in which multiple candidate tasks must be evaluated within a prescribed design domain. For a fair comparison, all methods are tested on the same unseen-task set with a fixed random seed; thus, the reported mean \(\pm\) SD summarizes task-wise variability within the benchmark rather than run-to-run variability. Second, repeated transfer analyses quantify run-to-run variability by averaging over five independent random seeds. Third, clustering stability is evaluated separately over 20 \(k\)-means seeds. To further verify that the fixed-benchmark results are not driven by a particular seed, representative 10-task evaluations were additionally repeated over 10 independent runs with different random seeds while keeping the unseen-task sets unchanged. The same comparative ranking was preserved, and the corresponding 95\% confidence intervals supported the same conclusion. Detailed seed-sensitivity results and the additional uncertainty analysis for representative fixed-benchmark reduction rates are reported in Appendix~C.}

% ===============================Table 1: Start==========================================
\begin{table*}[t] 
\singlespacing 
\centering 
\setcounter{table}{0}
% ▼▼▼ [수정 1] "Table 1:" 라벨 자체를 파란색으로 변경하는 명령어 추가 ▼▼▼
% \captionsetup{labelfont={color=blue}}

\caption{\hl{Comparison of average MSE and standard deviation (SD) on 10 unseen tasks: Group~A denotes the top-5 tasks with the highest learning-affinity metrics, while Group~B denotes the remaining bottom-5 tasks. Statistical significance ($p < 0.05$) was verified via two-sided paired Wilcoxon signed-rank tests.}} 
\label{tab:mse_comparison} 
 
% ===== global font size & spacing (Modified: Increased by 1pt) ===== 
{\fontsize{8.5pt}{10.5pt}\selectfont  % ★ 7.5pt -> 8.5pt
\renewcommand{\arraystretch}{1.25}    % ★ 1.35 -> 1.25 (Balance adjustment)
\setlength{\tabcolsep}{5pt}           % ★ 6pt -> 5pt (Prevent overflow)
% \color{blue}
% 그룹 블록(Helmholtz/Burgers/Linear) 사이 기본 간격
\newcommand{\GroupGap}{1.5pt} 
 
% 경계 '값'들만 더 벌리고 싶을 때 쓰는 추가 간격
\newcommand{\DataGap}{10pt}              
\newcommand{\dg}[1]{\hspace{\DataGap}#1} 
 
% (1) 헤더 전용 폰트 (+1pt adjusted)
\newcommand{\HeaderInc}{\fontsize{9.5pt}{11.5pt}\selectfont} % ★ 8.5pt -> 9.5pt

% (1-1) 인덱스 헤더 폰트 (+1pt adjusted)
\newcommand{\IndexHead}[1]{{\fontsize{9pt}{10.8pt}\selectfont\textbf{#1}}} % ★ 8pt -> 9pt
 
% (2) mean ± std  +  지수(E+00) 공통 폰트
\newcommand{\subnum}[1]{{\fontsize{6.5pt}{7.5pt}\selectfont #1}}  % ± 오른쪽 / 지수에 공통 사용
\newcommand{\meansd}[2]{\ensuremath{#1\pm\text{\subnum{#2}}}}
\newcommand{\expnum}[2]{#1\text{\subnum{#2}}}                      % #1: mantissa, #2: E+00 등

% ==== “10 Tasks” 텍스트만 오른쪽으로 밀기 ====
\newlength{\IndexSpace}
\setlength{\IndexSpace}{0.4em}
\newcommand{\ShiftRight}[2][1]{\makebox[0pt][l]{\hspace*{#1\IndexSpace}#2}\phantom{#2}}
% ============================================================================

\resizebox{\textwidth}{!}{% 
\begin{tabular}{ 
  c 
  c@{\hspace{\GroupGap}} % Method와 Helmholtz 사이 간격 
  c c c                   % Helmholtz (3) 
  @{\hspace{\GroupGap}}   % Helmholtz ↔ Burgers 기본 간격 
  c c c                   % Burgers' (3) 
  @{\hspace{\GroupGap}}   % Burgers ↔ Linear 기본 간격 
  c c c                   % Linear (3) 
} 
\specialrule{1.2pt}{0pt}{1.2pt} 
 
% === (1) 상위 공통 인덱스: Average MSE === 
% ★ Method 셀을 3줄 병합해서 세로 가운데 정렬
\multirow{3}{*}{\IndexHead{Method}} & 
\multicolumn{1}{c}{} & 
\multicolumn{9}{c}{{\HeaderInc\textbf{Average MSE} (\(\pm\sigma\))}} \\ 
\cmidrule(lr){3-11} 
 
% === 방정식 이름 라인 === 
& \multicolumn{1}{c}{} & 
\multicolumn{3}{c}{{\HeaderInc\textbf{Helmholtz}}} & 
\multicolumn{3}{c}{{\HeaderInc\textbf{Burgers'}}} & 
\multicolumn{3}{c}{{\HeaderInc\textbf{Linear~Elasticity}}} \\ 
\cmidrule(lr){3-5}\cmidrule(lr){6-8}\cmidrule(lr){9-11} 
 
% === 인덱스(열 이름) === 
& & 
\IndexHead{10\,Tasks} & \IndexHead{A~Group} & \IndexHead{B~Group} & 
\IndexHead{\ShiftRight{10\,Tasks}} & \IndexHead{A~Group} & \IndexHead{B~Group} & 
\IndexHead{\ShiftRight{10\,Tasks}} & \IndexHead{A~Group} & \IndexHead{B~Group} \\ 
\cmidrule(l){1-1}\cmidrule(l){3-3}\cmidrule(l){4-4}\cmidrule(l){5-5} 
\cmidrule(l){6-6}\cmidrule(l){7-7}\cmidrule(l){8-8}\cmidrule(l){9-9}\cmidrule(l){10-10}\cmidrule(l){11-11} 
\textbf{PINN-Transfer \cite{raissi2019physics}} & & 
\meansd{\expnum{4.36}{E+00}}{5.63} & \meansd{\expnum{8.69}{E+00}}{4.95} & \meansd{\expnum{3.82}{E-02}}{0.08} & 
\dg{\meansd{\expnum{5.08}{E-01}}{0.097}} & \meansd{\expnum{5.78}{E-01}}{0.063} & \meansd{\expnum{4.38}{E-01}}{0.071} & 
\dg{\meansd{\expnum{3.54}{E-02}}{0.0284}} & \meansd{\expnum{5.85}{E-02}}{0.0193} & \meansd{\expnum{1.23}{E-02}}{0.0104} \\ 
 
\textbf{PINN-scratch \cite{raissi2019physics}} & & 
\meansd{\expnum{6.12}{E+00}}{7.19} & \meansd{\expnum{1.19}{E+01}}{5.69} & \meansd{\expnum{3.53}{E-01}}{0.78} & 
\dg{\meansd{\expnum{2.03}{E+00}}{1.116}} & \meansd{\expnum{2.96}{E+00}}{0.665} & \meansd{\expnum{1.11}{E+00}}{0.472} & 
\dg{\meansd{\expnum{5.96}{E-02}}{0.0576}} & \meansd{\expnum{1.08}{E-01}}{0.0322} & \meansd{\expnum{1.12}{E-02}}{0.0237} \\ 
 
\textbf{MAML \cite{finn2017model}} & & 
\meansd{\expnum{4.06}{E+00}}{5.15} & \meansd{\expnum{8.06}{E+00}}{4.43} & \meansd{\expnum{5.34}{E-02}}{0.12} & 
\dg{\meansd{\expnum{7.09}{E-01}}{0.094}} & \meansd{\expnum{7.83}{E-01}}{0.041} & \meansd{\expnum{6.35}{E-01}}{0.068} & 
\dg{\meansd{\expnum{2.91}{E-02}}{0.0242}} & \meansd{\expnum{4.92}{E-02}}{0.0146} & \meansd{\expnum{9.05}{E-03}}{0.0098} \\ 
 
\textbf{ConML \cite{wu2025learning}} & & 
\meansd{\expnum{1.69}{E+00}}{3.195} & \meansd{\expnum{3.35}{E+00}}{3.863} & \meansd{\expnum{3.80}{E-02}}{0.024} & 
\dg{\meansd{\expnum{1.33}{E-01}}{0.242}} & \meansd{\expnum{2.52}{E-01}}{0.298} & \meansd{\expnum{1.45}{E-02}}{0.023} & 
\dg{\meansd{\expnum{3.15}{E-02}}{0.0325}} & \meansd{\expnum{5.95}{E-02}}{0.0228} & \meansd{\expnum{3.44}{E-03}}{0.0045} \\ 
 
\textbf{MAD \cite{huang2022meta}} & & 
\meansd{\expnum{3.55}{E+00}}{4.54} & \meansd{\expnum{7.07}{E+00}}{3.91} & \meansd{\expnum{3.32}{E-02}}{0.07} & 
\dg{\meansd{\expnum{2.49}{E-01}}{0.294}} & \meansd{\expnum{4.83}{E-01}}{0.238} & \meansd{\expnum{1.39}{E-02}}{0.030} & 
\dg{\meansd{\expnum{2.75}{E-02}}{0.0158}} & \meansd{\expnum{4.08}{E-02}}{0.0087} & \meansd{\expnum{1.42}{E-02}}{0.0068} \\ 
 
% ▼▼▼ [수정 2] Hyper-LR-PINN 행: 각 셀마다 \hl{...} 적용 ▼▼▼
\hl{\textbf{Hyper-LR-PINN \cite{cho2023hypernetwork}}} & & 
\hl{\meansd{\expnum{1.86}{E+00}}{3.75}} & 
\hl{\meansd{\expnum{3.60}{E+00}}{4.90}} & 
\hl{\meansd{\expnum{1.22}{E-01}}{0.09}} & 
\dg{\hl{\meansd{\expnum{1.60}{E-01}}{0.248}}} & 
\hl{\meansd{\expnum{3.18}{E-01}}{0.277}} & 
\hl{\meansd{\expnum{2.42}{E-03}}{0.003}} & 
\dg{\hl{\meansd{\expnum{4.27}{E-03}}{0.0035}}} & 
\hl{\meansd{\expnum{6.91}{E-03}}{0.0026}} & 
\hl{\meansd{\expnum{1.63}{E-03}}{0.0017}} \\
 
% ▼▼▼ [수정 3] P2INN 행: 각 셀마다 \hl{...} 적용 ▼▼▼
\hl{\textbf{P$^2$INN \cite{cho2024ppinn}}} & & 
\hl{\meansd{\expnum{2.99}{E+00}}{5.734}} & 
\hl{\meansd{\expnum{5.85}{E+00}}{7.325}} & 
\hl{\meansd{\expnum{1.49}{E-01}}{0.162}} & 
\dg{\hl{\meansd{\expnum{2.20}{E+00}}{3.495}}} & 
\hl{\meansd{\expnum{4.36}{E+00}}{3.977}} & 
\hl{\meansd{\expnum{3.55}{E-02}}{0.056}} & 
\dg{\hl{\meansd{\expnum{1.18}{E-02}}{0.0145}}} & 
\hl{\meansd{\expnum{2.15}{E-02}}{0.0152}} & 
\hl{\meansd{\expnum{2.06}{E-03}}{0.0026}} \\

% ▼▼▼ [수정 4] DATS-w 행: 각 셀마다 \hl{...} 적용 ▼▼▼
\hl{\textbf{DATS-w \cite{toloubidokhti2024dats}}} & &
\hl{\meansd{\expnum{2.84}{E+00}}{3.553}} &
\hl{\meansd{\expnum{5.32}{E+00}}{3.585}} &
\hl{\meansd{\expnum{3.58}{E-01}}{0.380}} &
\dg{\hl{\meansd{\expnum{2.57}{E-01}}{0.501}}} &
\hl{\meansd{\expnum{5.09}{E-01}}{0.637}} &
\hl{\meansd{\expnum{3.95}{E-03}}{0.004}} &
\dg{\hl{\meansd{\expnum{4.12}{E-03}}{0.0030}}} &
\hl{\meansd{\expnum{6.48}{E-03}}{0.0024}} &
\hl{\meansd{\expnum{1.75}{E-03}}{0.0009}} \\

\textbf{LAM-PINN (ours)} & & 
{\bfseries\boldmath \meansd{\expnum{1.45}{E-01}}{0.15}} & 
{\bfseries\boldmath \meansd{\expnum{2.72}{E-01}}{0.10}} & 
{\bfseries\boldmath \meansd{\expnum{1.82}{E-02}}{0.04}} & 
\dg{{\bfseries\boldmath \meansd{\expnum{5.88}{E-02}}{0.072}}} & 
{\bfseries\boldmath \meansd{\expnum{1.16}{E-01}}{0.060}} & 
{\bfseries\boldmath \meansd{\expnum{1.97}{E-03}}{0.009}} & 
\dg{{\bfseries\boldmath \meansd{\expnum{1.14}{E-03}}{0.0021}}} & 
{\bfseries\boldmath \meansd{\expnum{1.58}{E-03}}{0.0017}} & 
{\bfseries\boldmath \meansd{\expnum{6.87}{E-04}}{0.0002}} \\

\specialrule{1.2pt}{0pt}{0pt} 
\end{tabular} 
} 
} % end font+spacing scope 
\vspace{0.2cm}
\end{table*}
% ===============================Table 1: End==========================================

\begin{figure}[t!]
\centering
\includegraphics[width=1\linewidth]{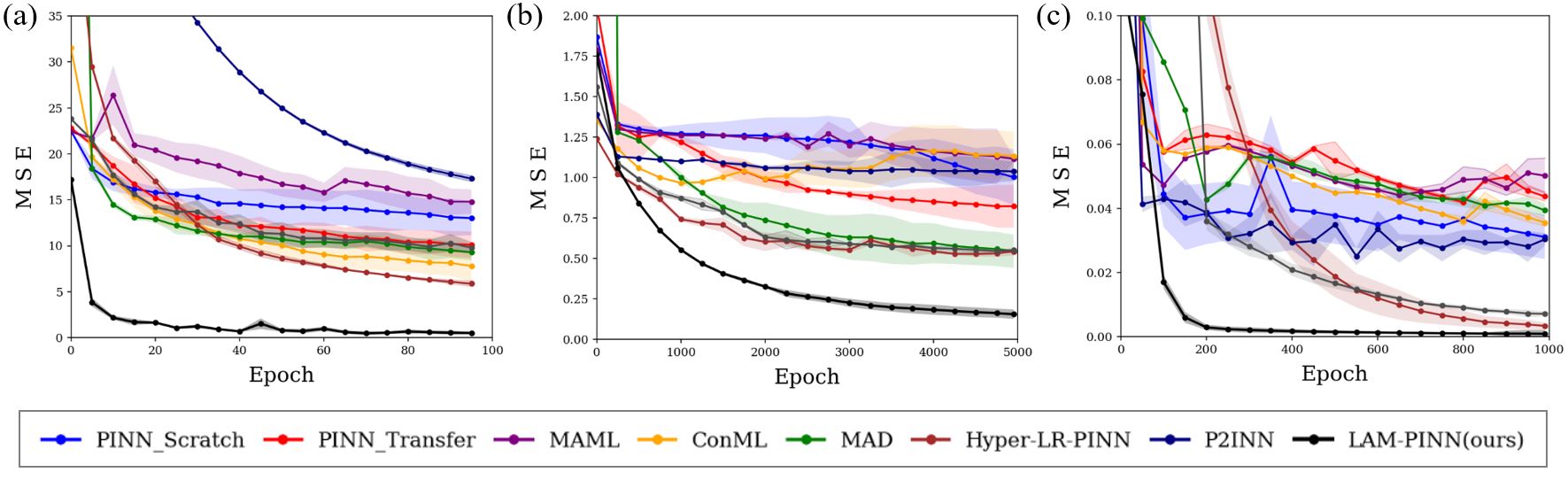}
% \captionsetup{labelfont={color=blue}, textfont={color=blue}}
\caption{
MSE convergence on unseen tasks across three PDE benchmarks: (a) Helmholtz, (b) Burgers’, and (c) Linear elasticity equation. Curves show the mean over 5 seeds; shaded bands indicate ± SD.
}
\label{fig:fig6}
\end{figure}

\subsection{Evaluation on PDEs}

\hl{We evaluate both LAM-PINN and baseline methods on three representative PDE families—Helmholtz, Burgers', and linear elasticity—under the DoE protocol described in Sec.~4.1. All methods were trained on the same task sets and compared through transfer to unseen tasks. Quantitative comparisons with the baselines are presented in Table~\ref{tab:mse_comparison}, transfer convergence curves are shown in Fig.~\ref{fig:fig6}, and representative transfer visualizations are provided in Fig.~\ref{fig:fig7}. Benchmark-specific PDE details are provided in Appendix~A, while the method-specific settings and computational budgets are summarized in Table~\ref{tab:baseline_configs}.}

Across all three PDE families, Table~\ref{tab:mse_comparison} shows that LAM-PINN consistently achieves the lowest average MSE. This efficiency is further supported by Fig.~\ref{fig:fig6}, which illustrates the fastest early-stage error reduction during transfer. Under the same transfer budget, LAM-PINN reconstructs target solution fields more accurately than the baselines across all benchmarks, as visualized in Fig.~\ref{fig:fig7}. These collective findings validate that learning-affinity-based clustering and adaptive module reuse effectively enhance transfer across tasks exhibiting potential heterogeneity within a bounded design space.

\begin{figure}[t!]
\centering
\includegraphics[width=1\linewidth]{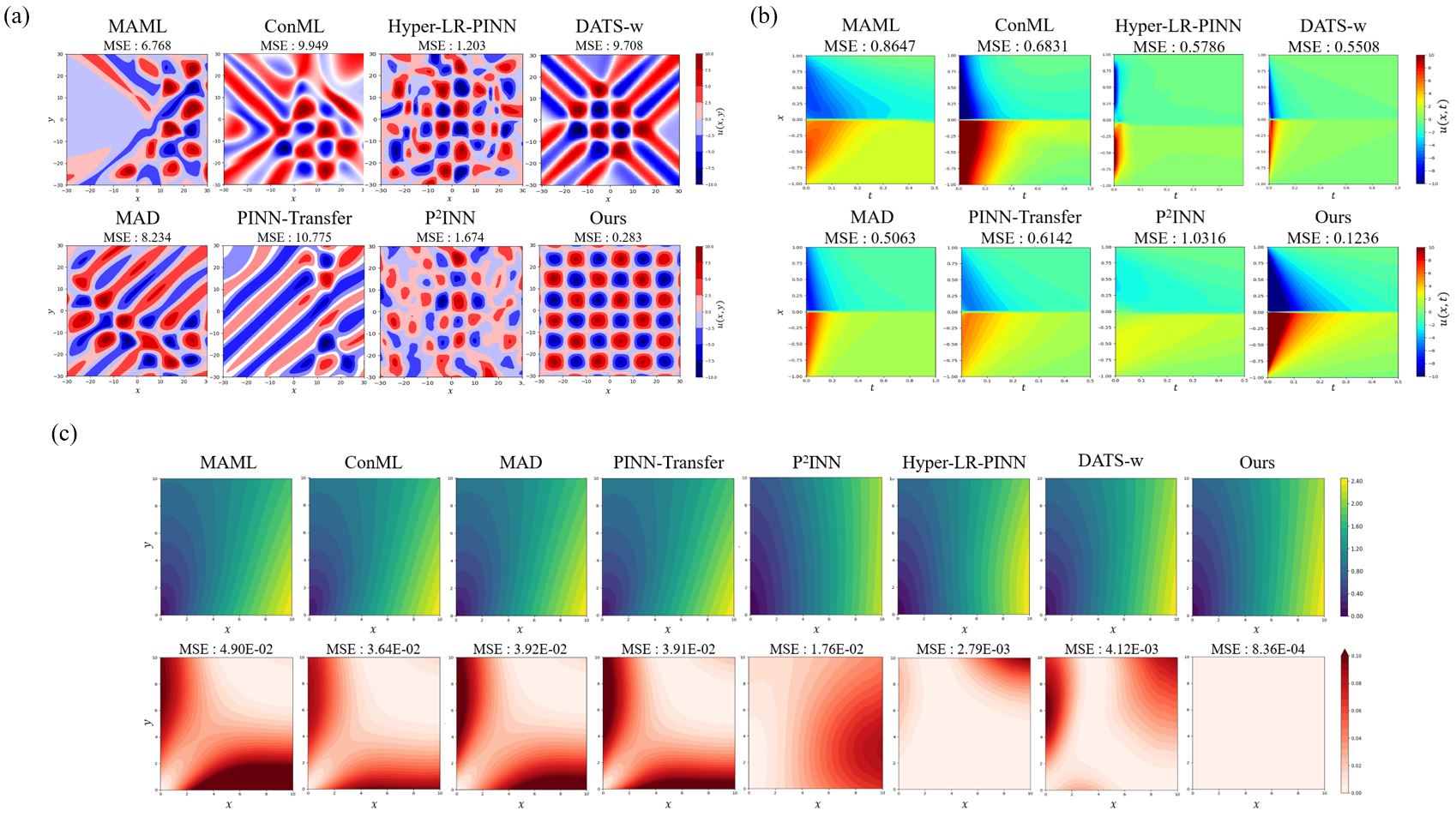}
% \captionsetup{labelfont={color=blue}, textfont={color=blue}}
\caption{
Representative transfer results on unseen tasks for (a) Helmholtz, (b) Burgers’, and (c) linear elasticity. For (c), the second row displays MSE maps on a common scale.
}
\label{fig:fig7}
\end{figure}

\subsection{Ablation study and analysis}

\subsubsection{Task representation with learning-affinity metrics}

\begin{figure}[t!]
\centering
\includegraphics[width=0.8\linewidth]{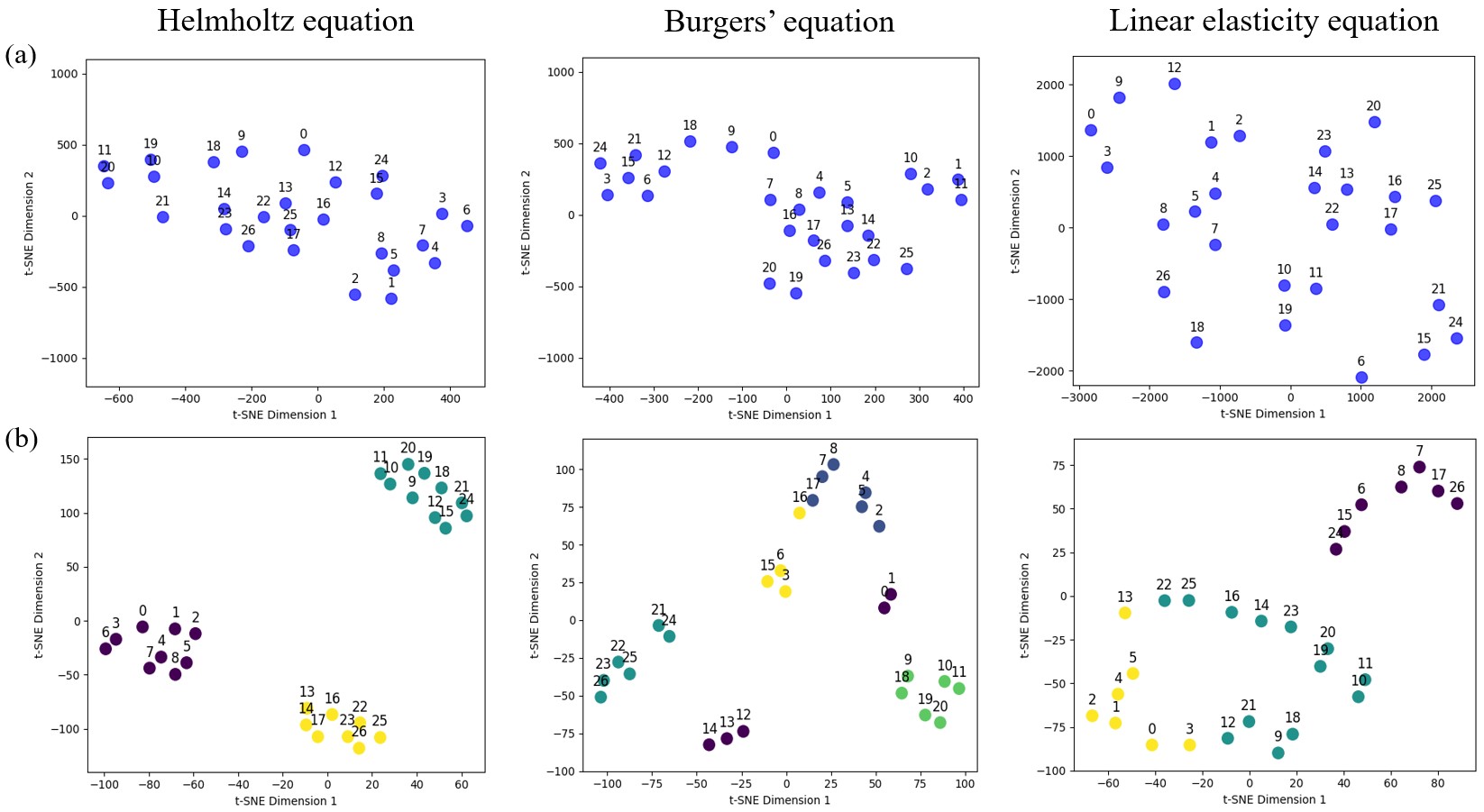}
\caption{
t-SNE visualization of task representations before and after applying the learning-affinity metrics for the three PDEs: (a) without the proposed metrics (based only on PDE parameters); (b) with the learning-affinity metrics incorporated.
}
\label{fig:fig10}
\end{figure}

To validate the efficacy of the proposed task representation, we first evaluate the specific contribution of learning-affinity metrics. For each of the three PDEs, we run brief transfer sessions using less than $5\%$ of the iterations required for a conventional PINN to converge, extract the three loss-based metrics, and concatenate them with the PDE parameters. The resulting embeddings are visualized using t-SNE \cite{van2008visualizing} in Fig.~\ref{fig:fig10}: using only PDE parameters (top row) yields entangled, weakly separable clusters, whereas incorporating learning-affinity metrics (bottom row) produces distinct and more compact regions in the embedding space.

\vspace{4pt}  % 두 그림 사이 공간 최소화

\begin{figure}[t!]
\centering
\includegraphics[width=1\linewidth]{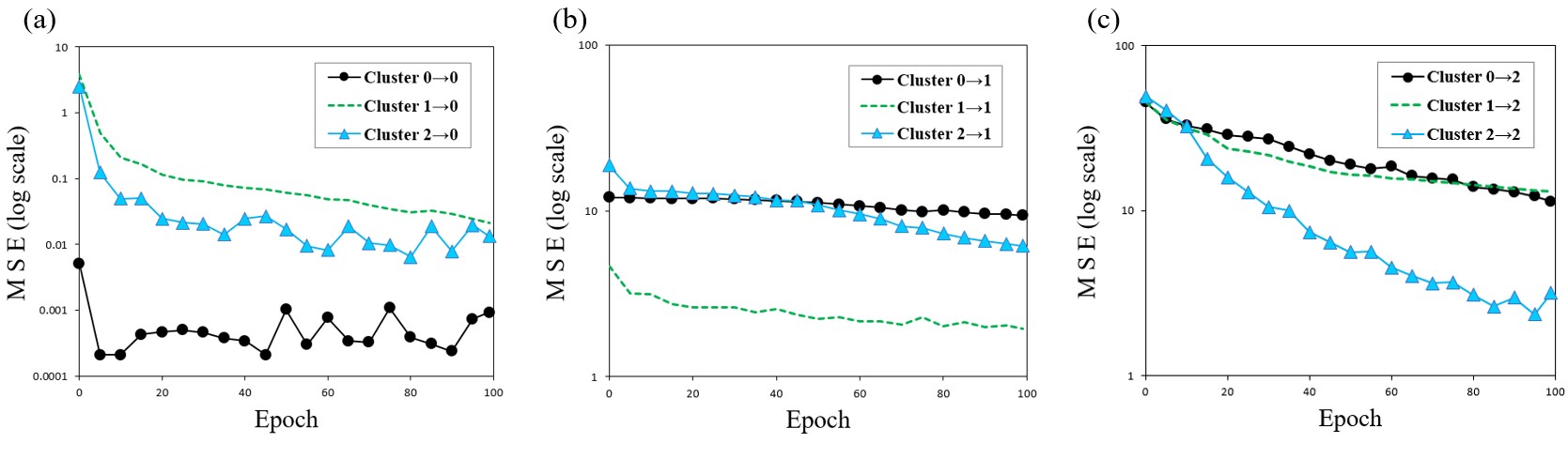}
\caption{
Validation of learning-affinity-based task clustering via transfer learning experiments on the Helmholtz equation. Each subfigure shows the epoch-wise MSE when transferring from a PINN pre-trained on a representative task (i.e., the one closest to the cluster centroid) to a target task sampled from a specific labeled cluster.  For example, "Cluster 0→1" denotes a transfer from a model pre-trained on a task near the center of cluster \#0 to a target task from cluster \#1.
}
\label{fig:fig11}
\end{figure}

To evaluate whether these clusters reflect meaningful learning behavior, we perform additional transfer experiments where a PINN pre-trained on a representative task (closest to each cluster centroid) is transferred either within the same cluster or across clusters. As shown in Fig.~\ref{fig:fig11}, within-cluster transfer consistently exhibits faster convergence and lower final MSE than cross-cluster transfer. Together with the group-wise results in Table~\ref{tab:mse_comparison}, this indicates that the proposed learning-affinity metrics capture both inter-task similarity and relative difficulty, providing a robust basis for structuring PINNs to handle task heterogeneity.

Table~\ref{tab:mse_comparison} presents a task-wise performance analysis on ten randomly sampled tasks for each PDE. For each case, two groups were compared: Group A comprises the five tasks with the highest learning-affinity metric values, whereas Group B includes the remaining five tasks with lower values. The proposed model outperformed all baselines in both groups, demonstrating robust performance even in Group A. Notably, while the baselines achieved comparable results in Group B, their performance degraded significantly in Group A. This indicates that the proposed learning-affinity metrics effectively reflect both task similarity and relative difficulty. Furthermore, the results highlight the limitations of conventional meta-learning approaches in handling wide task diversity and emphasize the utility of the proposed model in proactively addressing task heterogeneity through dynamic transfer guided by learning-affinity metrics.

% =============================== Table 2 (balanced & fits page width) ===============================
\begingroup
  % centered m-columns (horizontal & vertical centering)
  \newcolumntype{M}[1]{>{\centering\arraybackslash}m{#1}}

  % ---- fonts (numbers / exponent / std) ----
  \newcommand{\TblNumFont}{\fontsize{8pt}{9.6pt}\selectfont}   % mantissa
  \newcommand{\TblExpFont}{\fontsize{6pt}{7.2pt}\selectfont}   % exponent (E+00)
  \newcommand{\TblStdFont}{\fontsize{6pt}{7.2pt}\selectfont}   % ± std
  \newcommand{\HdrFont}{\fontsize{7pt}{8pt}\selectfont}
  \newcommand{\HdrSm}{\fontsize{6pt}{7pt}\selectfont}          % NEW: 1pt smaller for the 4 header cells only

  % mantissa + exponent (e.g., {1.85}{E+00})
  \newcommand{\Efmt}[2]{\text{\TblNumFont #1}\text{\TblExpFont #2}}

  % mean ± std (normal / bold)
  \newcommand{\MSEval}[2]{\ensuremath{\Efmt#1\,\text{\TblStdFont $\pm$ #2}}}
  \newcommand{\BMSEval}[2]{\ensuremath{\textbf{\Efmt#1}\,\text{\bfseries\TblStdFont $\pm$ #2}}}

  % aliases (기존 이름 유지)
  \newcommand{\TfourmseA}[2]{\MSEval{#1}{#2}}
  \newcommand{\TfourmseB}[2]{\MSEval{#1}{#2}}
  \newcommand{\TfourmseC}[2]{\MSEval{#1}{#2}}
  \newcommand{\TfourmseD}[2]{\MSEval{#1}{#2}}
  \newcommand{\TfourbmseA}[2]{\BMSEval{#1}{#2}}
  \newcommand{\TfourbmseB}[2]{\BMSEval{#1}{#2}}
  \newcommand{\TfourbmseC}[2]{\BMSEval{#1}{#2}}
  \newcommand{\TfourbmseD}[2]{\BMSEval{#1}{#2}}

  \begin{table*}[t]
    \singlespacing
    \centering

    % 캡션(라벨 + 텍스트) 파란색
    % \captionsetup{font=small, labelfont={bf,color=blue}, textfont={color=blue}}

    \caption{Ablation study on task representation and adaptive transfer across different configurations. For each PDE configuration, transfer learning is repeated over 5 seeds, and the average MSE $\pm$ SD is reported.}
    \label{tab:ablation_full}

    {\fontsize{8pt}{9.6pt}\selectfont
    \renewcommand{\arraystretch}{1.22}
    \setlength{\tabcolsep}{5.2pt}

    \resizebox{0.94\textwidth}{!}{%
    {
    \begin{tabular}{
      M{0.12\textwidth}
      M{0.07\textwidth} M{0.07\textwidth} M{0.07\textwidth}
      M{0.10\textwidth}
      M{0.14\textwidth} M{0.14\textwidth}
      M{0.14\textwidth} M{0.14\textwidth}
    }
    \specialrule{1.2pt}{0pt}{1.2pt}

    % Row-1 header
    \multicolumn{1}{c}{} &
    \multicolumn{4}{c}{\textbf{Model Configuration}} &
    \multicolumn{4}{c}{\textbf{MSE ($\pm \sigma$)}} \\

    % Row-2 header
    \cmidrule(lr){2-5}\cmidrule(lr){6-9}
    \textbf{Case} &
      \multicolumn{3}{c}{\shortstack{\HdrFont \textbf{Clustering}\\\HdrFont \textbf{Features}}} &
      \shortstack{\HdrFont \textbf{Adaptive}\\\HdrFont \textbf{Transfer}} &
      \multicolumn{2}{c}{\shortstack{\textbf{Helmholtz equation}\\\HdrFont with $(A,B,C)$}} &
      \multicolumn{2}{c}{\shortstack{\textbf{Burgers' equation}\\\HdrFont with $(\alpha,\nu,A)$}} \\

    % Row-3 header (ONLY here: 1pt smaller + slight LR tighten)
    \cmidrule(lr){2-4}\cmidrule(lr){5-5}\cmidrule(lr){6-7}\cmidrule(lr){8-9}
    &
      \shortstack{\HdrSm\hspace{-0.3pt}\textbf{PDE}\\\HdrSm\hspace{-0.3pt}\textbf{Parameters}} & % NEW
      \shortstack{\HdrSm\hspace{-0.3pt}\textbf{Random}\\\HdrSm\hspace{-0.3pt}\textbf{Metrics}} &  % NEW
      \shortstack{\HdrSm\hspace{-0.3pt}\textbf{Learning}\\\HdrSm\hspace{-0.3pt}\textbf{Affinity}\\\HdrSm\hspace{-0.3pt}\textbf{Metrics}} & % NEW
      \shortstack{\HdrSm\hspace{-0.3pt}\textbf{Learnable}\\\HdrSm\hspace{-0.3pt}$\boldsymbol{\lambda}$} & % NEW
      {\HdrFont (10, 3, 3)} & {\HdrFont (7, 2, 3.5)} &
      {\HdrFont (1.8, 0.006, 9)} & {\HdrFont (0.12, 0.045, 9.6)} \\

    \cmidrule(lr){1-1}\cmidrule(lr){2-4}\cmidrule(lr){5-5}\cmidrule(lr){6-7}\cmidrule(lr){8-9}

    % ================= Data =================
    \textbf{Case \#1}
      & $\checkmark$ &              &              & $\checkmark$
      & \TfourmseA{{1.85}{E+00}}{0.42}
      & \TfourmseB{{2.57}{E+00}}{0.55}
      & \TfourmseC{{6.38}{E-01}}{0.058}
      & \TfourmseD{{8.27}{E-03}}{1.3E-04} \\

    \textbf{Case \#2}
      & $\checkmark$ & $\checkmark$ &              & $\checkmark$
      & \TfourmseA{{2.80}{E+00}}{0.38}
      & \TfourmseB{{3.21}{E+00}}{0.33}
      & \TfourmseC{{5.95}{E-01}}{0.246}
      & \TfourmseD{{2.48}{E-03}}{5.0E-05} \\

    \textbf{Case \#3}
      & $\checkmark$ &              & $\checkmark$ &
      & \TfourmseA{{1.84}{E+00}}{0.31}
      & \TfourmseB{{2.45}{E+00}}{0.44}
      & \TfourmseC{{4.82}{E-01}}{0.059}
      & \TfourmseD{{7.11}{E-03}}{9.3E-04} \\

    \textbf{LAM-PINN (ours)}
      & $\checkmark$ &              & $\checkmark$ & $\checkmark$
      & \TfourbmseA{{3.07}{E-01}}{0.12}
      & \TfourbmseB{{3.31}{E-01}}{0.17}
      & \TfourbmseC{{1.04}{E-01}}{0.023}
      & \TfourbmseD{{1.60}{E-04}}{9.0E-05} \\

    \specialrule{1.2pt}{0pt}{0pt}
    \end{tabular}
    }% end \color{blue}
    }% end resizebox
    }% end fontsize
  \end{table*}

\endgroup
% =============================== End Table 2 =========================================================
\begin{figure}[t!]
\centering
\includegraphics[width=1\linewidth]{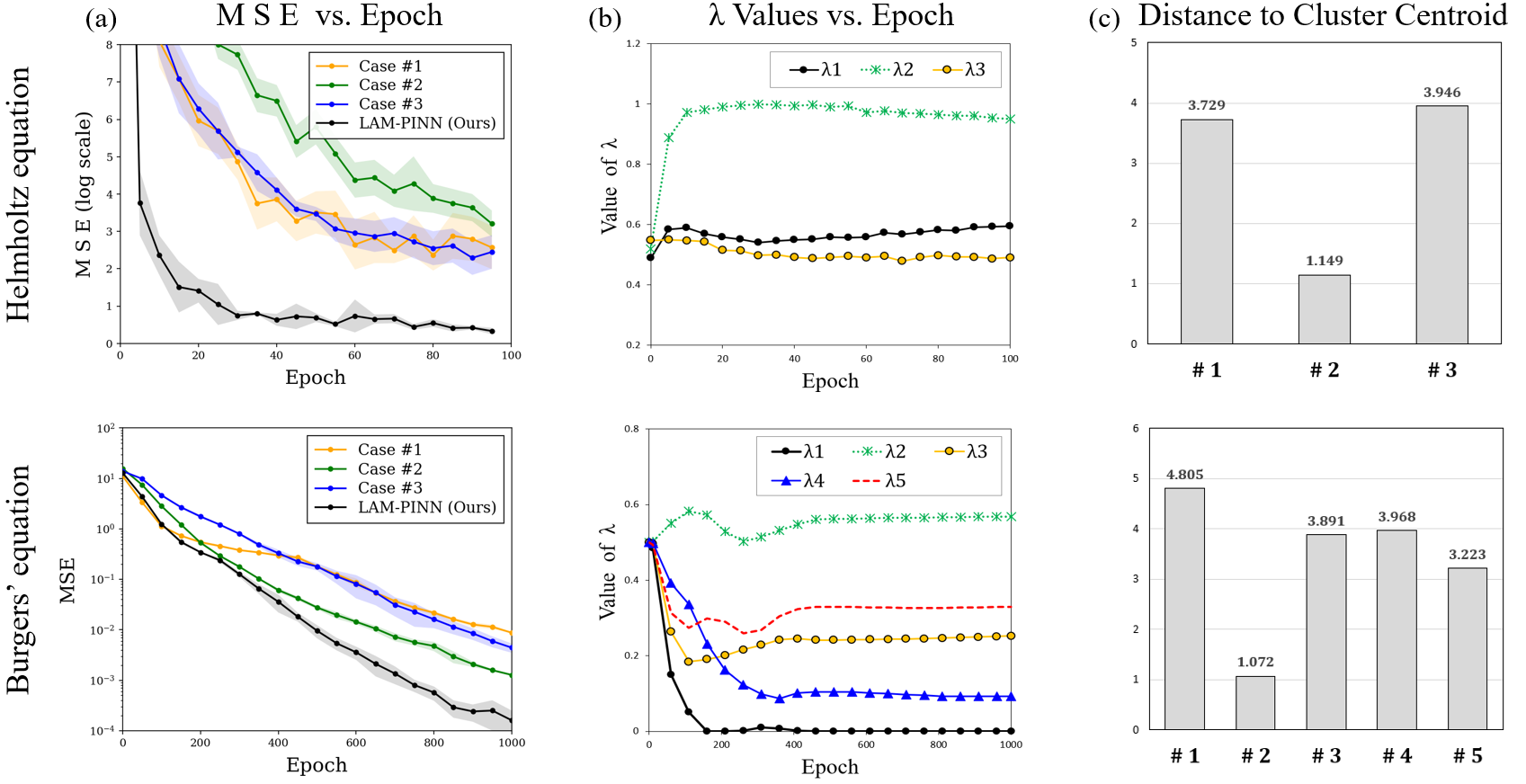}
\caption{
(a) MSE over training epochs for an unseen task under each ablation setting. Curves show the average across 5 seeds; shaded bands indicate ± SD. (b) Evolution of learnable parameters $\lambda_j$ during transfer, showing adaptive subnetwork contributions. (c) Euclidean distances from the unseen task to each cluster centroid in the task representation space constructed using learning-affinity metrics.
}
\label{fig:fig12}
\end{figure}

\subsubsection{Learning-affinity metrics for effective selective transfer}

To evaluate the combined impact of learning-affinity-based clustering and $\lambda$-weighted subnetworks on adaptive transfer, we conducted an ablation study. Table~\ref{tab:ablation_full} compares four configurations: (Case~\#1) clustering using only PDE parameters; (Case~\#2) clustering with PDE parameters plus random auxiliary metrics; (Case~\#3) clustering using the proposed learning-affinity metrics but with fixed $\lambda$; and the full LAM-PINN equipped with both learning-affinity clustering and learnable $\lambda$.

Across unseen tasks in both Helmholtz and Burgers' equations, LAM-PINN achieves the lowest MSE and the fastest error decay (Fig.~\ref{fig:fig12}(a)). Using random metrics (Case~\#2) does not improve over parameter-only clustering, and disabling $\lambda$-adaptivity (Case~\#3) substantially degrades performance even when the clusters are well structured. Fig.~\ref{fig:fig12}(b) demonstrates that the learned $\lambda_j$ rapidly converge to prioritize a specific subnetwork. Crucially, Fig.~\ref{fig:fig12}(c) validates this adaptive selection: it demonstrates that the subnetwork receiving the dominant weight ($\lambda_j$) aligns with the cluster centroid closest to the unseen task. These results highlight that learning-affinity-driven clustering and $\lambda$-adaptive routing are complementary and jointly necessary for robust selective transfer.

% ===============================Table 3: Start==========================================

\begin{table*}[t]
% \color{blue}
\singlespacing
\centering
% Requires: \usepackage{booktabs, array, multirow, caption}
% \captionsetup[table]{labelfont={normalfont, color=blue}, textfont={normalfont, color=blue}}
% \captionsetup{labelfont={color=blue}, textfont={color=blue}}
\setcounter{table}{2}  % 다음 \caption 이 Table 3으로 출력됨

\caption{
% \color{blue}
DoE robustness on the Helmholtz equation: runtime and average MSE (\(\pm\sigma\)) across alternative designs---Random 27 tasks, (\(2\times2\times2\)), (\(3\times3\times2\)), (\(3\times3\times3\)), and (\(4\times4\times3\)). Group~A denotes the top-5 tasks with the highest learning-affinity metrics, while Group~B denotes the remaining bottom-5 tasks. Statistical significance ($p < 0.05$) was verified via two-sided paired Wilcoxon signed-rank tests.
}
\label{tab:doe_robustness_helmholtz}

\newcolumntype{M}[1]{>{\centering\arraybackslash}m{#1}}
\newcommand{\meansd}[2]{\ensuremath{#1\pm{\fontsize{6pt}{7.2pt}\selectfont #2}}}

{\fontsize{7pt}{8.4pt}\selectfont
\renewcommand{\arraystretch}{1.15}
\setlength{\tabcolsep}{5pt}

\resizebox{0.95\textwidth}{!}{%
\begin{tabular}{M{0.10\textwidth} M{0.20\textwidth} M{0.11\textwidth} M{0.16\textwidth} c c c}
\specialrule{1.2pt}{0pt}{1.2pt}

\multirow{2}{*}[-0.2ex]{\textbf{Case}} &
\multirow{2}{*}[-0.2ex]{\textbf{DoE setting}} &
\multirow{2}{*}[-0.2ex]{\shortstack{\textbf{No. of}\\\textbf{tasks}}} &
\multirow{2}{*}[-0.2ex]{\shortstack{\textbf{Time}\\\textbf{(preproc/train)}\\\textbf{[min]}}} &
\multicolumn{3}{c}{\textbf{Average MSE} (\(\pm\sigma\))} \\
\cmidrule(lr){5-7}
& & & & \textbf{10 Tasks} & \textbf{Group A} & \textbf{Group B} \\

\cmidrule(lr){1-1}\cmidrule(lr){2-2}\cmidrule(lr){3-3}\cmidrule(lr){4-4}\cmidrule(lr){5-5}\cmidrule(lr){6-6}\cmidrule(lr){7-7}

Case~\#1 & Random (27 tasks) & 27 & 11.5 / 99 &
\meansd{1.16\text{\tiny{E+00}}}{2.211} &
\meansd{2.25\text{\tiny{E+00}}}{2.829} &
\meansd{6.32\text{\tiny{E-02}}}{0.048} \\

Case~\#2 & \(2\times2\times2\) & 8 & 3.5 / 29.5 &
\meansd{1.15\text{\tiny{E+00}}}{2.146} &
\meansd{2.25\text{\tiny{E+00}}}{2.709} &
\meansd{5.13\text{\tiny{E-02}}}{0.051} \\

Case~\#3 & \(3\times3\times2\) & 18 & 8.5 / 69 &
\meansd{1.04\text{\tiny{E+00}}}{2.089} &
\meansd{2.05\text{\tiny{E+00}}}{2.704} &
\meansd{4.36\text{\tiny{E-02}}}{0.030} \\

\textbf{Case~\#4} &
\textbf{\boldmath \(3\times3\times3\) (Ours)} &
\textbf{27} &
\textbf{11.5 / 99} &
\bfseries\boldmath \meansd{1.45\text{\tiny{E-01}}}{0.152} &
\bfseries\boldmath \meansd{2.72\text{\tiny{E-01}}}{0.102} &
\bfseries\boldmath \meansd{1.82\text{\tiny{E-02}}}{0.040} \\

Case~\#5 & \(4\times4\times3\) & 48 & 20.5 / 178.5 &
\meansd{1.21\text{\tiny{E-01}}}{0.186} &
\meansd{2.27\text{\tiny{E-01}}}{0.223} &
\meansd{1.44\text{\tiny{E-02}}}{0.007} \\

\specialrule{1.2pt}{0pt}{0pt}
\end{tabular}
} % end resizebox
} % end fontsize group
\vspace{0.3cm}
\end{table*}

% ===============================Table 3: End==========================================

\subsection{Further analysis}

\subsubsection{Analysis of DoE rationale and clustering stability}

To justify the $3\times3\times3$ full-factorial DoE choice and examine scalability, we evaluate five task sets, including a denser $4\times4\times3$ design with 48 tasks. As shown in Table 3, Case~\#4 achieves an average MSE nearly an order of magnitude lower than the lower-level designs (Cases~\#1–\#3), particularly in Group A. Although Case~\#5 remains numerically stable and slightly improves MSE, it increases total preprocessing and training time by approximately 80\% versus Case~\#4, indicating diminishing returns and supporting $3\times3\times3$ as a practical cost--accuracy compromise~\cite{heckert2002handbook}. \hl{This efficiency-driven approach is reflected in our selection of the number of clusters $K$. Fig.~\ref{fig:fig.B.1} in Appendix B shows that a larger $K$ does not necessarily improve transfer performance but increases computational cost, validating $K=3$ as the optimal balance.}

In addition to this runtime trade-off, scalability in model capacity is tied to the modular nature of our architecture (Sec.~3.3). Unlike approaches that scale with the number of tasks, LAM-PINN learns cluster-level modules where model capacity is governed by task heterogeneity (the number of clusters $K$) rather than the raw task count. Consequently, additional tasks within a designed range primarily serve as supplementary training samples that reinforce existing specialized modules, ensuring that parameter growth remains moderate even as the task pool expands.

We also examine the stability of the learned task clusters. For each PDE and $K\in\{2,\dots,6\}$, we run $k$-means over 20 random seeds and report the mean silhouette score and average pairwise Adjusted Rand Index (ARI) (Table~\ref{tab:clustering_stats}). The selected $K$ values yield both good separation and high seed-level stability. Finally, to test the reliability of short transfer sessions, we repeat clustering on 2D and 3D Helmholtz equations using budgets of 1--20\% of full-convergence iterations. Table~\ref{tab:helmholtz_combined} shows that the preferred $K$ remains stable and that the label-disagreement rate drops to $\sim4\%$ at 4\% budget and reaches 0\% at 10\% in our experiments. We compute the label-disagreement rate by first aligning cluster indices between the short-session and full-epoch solutions using Hungarian matching and then measuring the fraction of tasks whose aligned assignments differ. These results support the use of a brief (less than 5\%) transfer budget as a practical default in resource-constrained environments.

% =============================== Combined Table: 2D (top) + 3D (bottom) ===============================
\begin{table*}[t]
\singlespacing
\centering
% \captionsetup{labelfont={bf,color=blue}, textfont={color=blue}}
\caption{Clustering quality for 2D (top) and 3D (bottom) Helmholtz tasks at different cluster numbers $K$ and training budgets. Entries are the mean silhouette and ARI over 20 seeds, and the last row reports the label disagreement rate (\%) with respect to the clustering at full convergence. $K$ is selected by jointly considering separation and stability to avoid bias toward smaller clusters (details in Appendix~B).}
\label{tab:helmholtz_combined}

\newcolumntype{M}[1]{>{\centering\arraybackslash}m{#1}}

% --- 전체 테이블 기본 폰트/행간/열 간격을 조금 더 줄임 ---
{\fontsize{6.5pt}{7.6pt}\selectfont   % 7pt -> 6.5pt
\renewcommand{\arraystretch}{0.98}    % 1.06 -> 0.98 (행간 살짝 축소)
\setlength{\tabcolsep}{3pt}           % 4pt -> 3pt (열 간격 축소)
% \color{blue}
% --- 폭도 줄여서 세로 크기까지 함께 축소 (0.98 -> 0.90) ---
\resizebox{0.90\textwidth}{!}{%
\begin{tabular}{c *{10}{c}}
\specialrule{1.2pt}{0pt}{1.0pt}

% ========================= 2D Helmholtz (TOP) =========================
\multicolumn{11}{c}{\textbf{2D Helmholtz}}\\
\addlinespace[1pt]   % 2pt -> 1pt

\multirow{2}{*}{\shortstack[c]{\textbf{Number}\\\textbf{of}\\\textbf{Cluster $K$}}} &
\multicolumn{2}{c}{\textbf{50 epochs (1\%)}} &
\multicolumn{2}{c}{\textbf{100 epochs (2\%)}} &
\multicolumn{2}{c}{\textbf{200 epochs (4\%)}} &
\multicolumn{2}{c}{\textbf{500 epochs (10\%)}} &
\multicolumn{2}{c}{\textbf{1000 epochs (20\%)}} \\
\cmidrule(lr){2-3}\cmidrule(lr){4-5}\cmidrule(lr){6-7}\cmidrule(lr){8-9}\cmidrule(lr){10-11}
& \textbf{Silhouette} & \textbf{ARI}
& \textbf{Silhouette} & \textbf{ARI}
& \textbf{Silhouette} & \textbf{ARI}
& \textbf{Silhouette} & \textbf{ARI}
& \textbf{Silhouette} & \textbf{ARI} \\
\cmidrule(lr){1-1}\cmidrule(lr){2-2}\cmidrule(lr){3-3}
\cmidrule(lr){4-4}\cmidrule(lr){5-5}\cmidrule(lr){6-6}\cmidrule(lr){7-7}
\cmidrule(lr){8-8}\cmidrule(lr){9-9}\cmidrule(lr){10-10}\cmidrule(lr){11-11}

% -- data rows (2D) --
2 & \msd{0.391}{0.003} & 1.00 & \msd{0.364}{0.004} & 0.98 & \msd{0.366}{0.020} & 1.00 & \msd{0.346}{0.032} & 0.94 & \msd{0.363}{0.025} & 0.97 \\
\textbf{3} & \msdb{0.334}{0.014} & \textbf{1.00} & \msdb{0.312}{0.002} & \textbf{1.00} & \msdb{0.308}{0.007} & \textbf{1.00} & \msdb{0.298}{0.016} & \textbf{1.00} & \msdb{0.292}{0.012} & \textbf{1.00} \\
4 & \msd{0.327}{0.011} & 0.95 & \msd{0.291}{0.022} & 0.96 & \msd{0.278}{0.024} & 0.89 & \msd{0.281}{0.020} & 0.90 & \msd{0.288}{0.021} & 0.89 \\
5 & \msd{0.326}{0.031} & 0.90 & \msd{0.309}{0.030} & 0.86 & \msd{0.322}{0.041} & 0.81 & \msd{0.283}{0.033} & 0.85 & \msd{0.211}{0.019} & 0.87 \\
6 & \msd{0.349}{0.024} & 0.87 & \msd{0.297}{0.021} & 0.78 & \msd{0.324}{0.022} & 0.78 & \msd{0.324}{0.034} & 0.83 & \msd{0.282}{0.030} & 0.82 \\

\addlinespace[0.5pt]
\cmidrule(lr){1-1}\cmidrule(lr){2-11}
\multicolumn{1}{M{0.10\textwidth}}{\textit{\shortstack{disagreement\\rate [\%]}}} &
\multicolumn{2}{M{0.18\textwidth}}{\msd{7.78}{1.14}\,\%} &
\multicolumn{2}{M{0.18\textwidth}}{\msd{0.37}{1.66}\,\%} &
\multicolumn{2}{M{0.18\textwidth}}{\msd{0.00}{0.00}\,\%} &
\multicolumn{2}{M{0.18\textwidth}}{\msd{0.00}{0.00}\,\%} &
\multicolumn{2}{M{0.18\textwidth}}{\msd{0.00}{0.00}\,\%} \\

\addlinespace[2pt]
\specialrule{0.8pt}{0pt}{2pt}

% ========================= 3D Helmholtz (BOTTOM) =========================
\multicolumn{11}{c}{\textbf{3D Helmholtz}}\\
\addlinespace[1pt]

\multirow{2}{*}{\shortstack[c]{\textbf{Number}\\\textbf{of}\\\textbf{Cluster $K$}}} &
\multicolumn{2}{c}{\textbf{100 epochs (1\%)}} &
\multicolumn{2}{c}{\textbf{200 epochs (2\%)}} &
\multicolumn{2}{c}{\textbf{400 epochs (4\%)}} &
\multicolumn{2}{c}{\textbf{1000 epochs (10\%)}} &
\multicolumn{2}{c}{\textbf{2000 epochs (20\%)}} \\
\cmidrule(lr){2-3}\cmidrule(lr){4-5}\cmidrule(lr){6-7}\cmidrule(lr){8-9}\cmidrule(lr){10-11}
& \textbf{Silhouette} & \textbf{ARI}
& \textbf{Silhouette} & \textbf{ARI}
& \textbf{Silhouette} & \textbf{ARI}
& \textbf{Silhouette} & \textbf{ARI}
& \textbf{Silhouette} & \textbf{ARI} \\
\cmidrule(lr){1-1}\cmidrule(lr){2-2}\cmidrule(lr){3-3}
\cmidrule(lr){4-4}\cmidrule(lr){5-5}\cmidrule(lr){6-6}\cmidrule(lr){7-7}
\cmidrule(lr){8-8}\cmidrule(lr){9-9}\cmidrule(lr){10-10}\cmidrule(lr){11-11}

% -- data rows (3D) --
\textbf{2} & \msdb{0.295}{0.005} & \textbf{0.97} & \msdb{0.304}{0.003} & \textbf{0.96} & \msdb{0.313}{0.007} & \textbf{0.94} & \msdb{0.315}{0.009} & \textbf{0.97} & \msdb{0.323}{0.008} & \textbf{1.00} \\
3 & \msd{0.242}{0.033} & 0.65 & \msd{0.249}{0.033} & 0.61 & \msd{0.251}{0.031} & 0.89 & \msd{0.257}{0.025} & 0.92 & \msd{0.259}{0.028} & 0.93 \\
4 & \msd{0.231}{0.023} & 0.77 & \msd{0.237}{0.021} & 0.73 & \msd{0.245}{0.018} & 0.75 & \msd{0.246}{0.016} & 0.74 & \msd{0.239}{0.023} & 0.67 \\
5 & \msd{0.211}{0.017} & 0.65 & \msd{0.217}{0.015} & 0.58 & \msd{0.220}{0.019} & 0.59 & \msd{0.227}{0.011} & 0.55 & \msd{0.219}{0.023} & 0.58 \\
6 & \msd{0.209}{0.012} & 0.54 & \msd{0.214}{0.012} & 0.59 & \msd{0.224}{0.020} & 0.61 & \msd{0.219}{0.022} & 0.62 & \msd{0.218}{0.021} & 0.64 \\

\addlinespace[0.5pt]
\cmidrule(lr){1-1}\cmidrule(lr){2-11}
\multicolumn{1}{M{0.10\textwidth}}{\textit{\shortstack{disagreement\\rate [\%]}}} &
\multicolumn{2}{M{0.18\textwidth}}{\msd{6.91}{1.87}\,\%} &
\multicolumn{2}{M{0.18\textwidth}}{\msd{6.91}{1.87}\,\%} &
\multicolumn{2}{M{0.18\textwidth}}{\msd{4.20}{1.87}\,\%} &
\multicolumn{2}{M{0.18\textwidth}}{\msd{0.00}{0.00}\,\%} &
\multicolumn{2}{M{0.18\textwidth}}{\msd{0.00}{0.00}\,\%} \\

\specialrule{1.2pt}{0pt}{0pt}
\end{tabular}
}% end resizebox
}% end fontsize
\end{table*}
% =============================== End Combined Table =========================================

\subsubsection{Feasibility in high-dimensional and irregular scenarios}
To assess the extensibility of LAM-PINN to complex engineering scenarios beyond the main 2D benchmarks, we apply it to two additional, more challenging cases: a 3D Helmholtz problem and a linear elasticity problem in a plate with a circular hole. Appendix~A provides the 3D Helmholtz configuration and the modified boundary conditions for the linear elasticity problem, which inherits its DoE ranges and PDE formulation from the main study. We compare LAM-PINN against conventional PINN strategies and Hyper-LR-PINN—one of the strongest baselines in the main benchmarks—while using only approximately 15\% of the pre-training budget allocated to the baselines. Under identical transfer budgets, LAM-PINN achieves an average MSE reduction of 87\% on the 3D Helmholtz problem and 45\% on the plate-with-hole problem, relative to the average performance of the baselines (Table~\ref{tab:feasibility_cases}). These results indicate that learning-affinity-based modularization remains effective in higher-dimensional and geometrically irregular domains, thereby supporting its potential for broader engineering applications.

% =============================== Table 5 (Feasibility on Challenging Cases) ===============================
\begingroup
\renewcommand{\thetable}{5} % force table number if needed
\begin{table*}[t]
\singlespacing
\centering
% \captionsetup{labelfont={bf,color=blue}, textfont={color=blue}}

% Caption Recommendation applied
\caption{Transfer performance on the 3D Helmholtz equation and linear elasticity with a circular hole. Epochs indicate the number of transfer learning iterations. Results are reported as average MSE $\pm$ SD over 5 seeds.}
\label{tab:feasibility_cases}
% \color{blue}
% column types
\newcolumntype{L}[1]{>{\raggedright\arraybackslash}m{#1}} % left-aligned, vertically centered
\newcolumntype{Y}[1]{>{\centering\arraybackslash}m{#1}}   % centered, vertically centered

% mean ± std macro: numbers are text-mode to fix italics and hyphen size
\newcommand{\msepm}[2]{#1\,\ensuremath{\pm}\,{\fontsize{6pt}{7pt}\selectfont #2}}
\newcommand{\msepmB}[2]{\textbf{#1}\,\ensuremath{\pm}\,{\fontsize{6pt}{7pt}\selectfont \textbf{#2}}}

{\fontsize{8pt}{9.3pt}\selectfont
\renewcommand{\arraystretch}{1.18}
\setlength{\tabcolsep}{5.0pt}

\resizebox{0.98\textwidth}{!}{%
\begin{tabular}{
  Y{0.202\textwidth}  % Method: Center-Center alignment
  Y{0.185\textwidth}  % 3D: 500
  Y{0.185\textwidth}  % 3D: 1000
  Y{0.185\textwidth}  % Plate: 1000
  Y{0.185\textwidth}  % Plate: 2000
}
\specialrule{1.2pt}{0pt}{1.2pt}

% Header Row 1: Method merged, Equations spanned
\multirow{2}{*}{\textbf{Method}} & \multicolumn{2}{c}{\textbf{3D Helmholtz Equation}} & \multicolumn{2}{c}{\textbf{Linear Elasticity: Plate with a Circular Hole}}\\
\cmidrule(lr){2-3}\cmidrule(lr){4-5}

% Header Row 2: Epochs (Method cell is skipped here due to multirow)
 & \textbf{500 epochs} & \textbf{1000 epochs} & \textbf{1000 epochs} & \textbf{2000 epochs} \\
\cmidrule(lr){1-1}\cmidrule(lr){2-5}

PINN-scratch \cite{raissi2019physics}
  & \msepm{1.53E-01}{0.0587}
  & \msepm{5.61E-02}{0.0088}
  & \msepm{1.73E-01}{0.0178}
  & \msepm{1.49E-01}{0.0165}
\\
PINN-Transfer \cite{raissi2019physics}
  & \msepm{1.77E-01}{0.0116}
  & \msepm{7.05E-02}{0.0032}
  & \msepm{1.33E-01}{0.0116}
  & \msepm{1.07E-01}{0.0153}
\\
Hyper-LR-PINN \cite{cho2023hypernetwork}
  & \msepm{1.69E-01}{0.0007}
  & \msepm{1.59E-02}{0.0029}
  & \msepm{1.64E-01}{0.0003}
  & \msepm{1.38E-01}{0.0008}
\\
% LAM-PINN Row: Font size reset to match baselines (8pt), bold kept
\textbf{LAM-PINN (ours)}
  & \msepmB{1.04E-02}{0.0027}
  & \msepmB{4.03E-03}{0.0009}
  & \msepmB{1.17E-01}{0.0100}
  & \msepmB{7.03E-02}{0.0033}
\\
\specialrule{1.2pt}{0pt}{0pt}
\end{tabular}
} % end resizebox
} % end fontsize group

\end{table*}
\endgroup
% =============================== End of Table 5 ============================================

\subsubsection{Comparison with advanced PINN generalization methods}

While the superior generalization performance of LAM-PINN has been established in the previous sections, its practical utility also depends on computational efficiency and model complexity. To explicitly evaluate these aspects, we compare LAM-PINN with two representative auxiliary-network-based schemes, Hyper-LR-PINN and P$^{2}$INN, on the linear elasticity benchmark (Table~\ref{tab:baseline_elasticity_refined}). Hyper-LR-PINN and P$^{2}$INN augment a base PINN with a hypernetwork or parameter encoder to map PDE parameters to task-conditioned latent vectors or weight coefficients. In our experiments, these baselines were allocated 20,000 training epochs per task to ensure convergence; notably, reducing this budget to 10,000 epochs leads to a marked performance drop, indicating their reliance on extensive training. In contrast, LAM-PINN uses 30{,}021 trainable parameters---comparable to Hyper-LR-PINN and only about 24\% of P$^{2}$INN---and converges in approximately 400 epochs per task, including the preprocessing step. Despite this approximately 50-fold smaller computational budget, the average MSE over ten unseen tasks is reduced by 73.3\% relative to the best-performing baseline (Hyper-LR-PINN at 20k epochs) and by 90.3\% relative to P$^{2}$INN. These results indicate that the proposed clustering and module selection strategy provides a structurally simpler and substantially more efficient route to generalization across heterogeneous PINN tasks than auxiliary-network-based generalization schemes.

% =============================== Table 6 (final: no-overlap, centered header, compact) ===============================
\begingroup
\renewcommand{\thetable}{6}
\begin{table*}[t]
\centering
% (세로 간격만) 캡션-테이블 간격 축소
% \captionsetup{labelfont={bf,color=blue}, textfont={color=blue}}
\caption{Efficiency comparison based on linear elasticity equation: MSE is the average $\pm$ SD for 10 unseen tasks. Statistical significance ($p < 0.05$) was verified via two-sided paired Wilcoxon signed-rank tests.}
\label{tab:baseline_elasticity_refined}
% \color{blue}

% (세로 간격만) booktabs 규칙선 위/아래 여백 축소
\setlength{\aboverulesep}{0.35ex}
\setlength{\belowrulesep}{0.35ex}
\setlength{\cmidrulesep}{0.20ex}

% mean ± sd macro  (SD 글자 크기만 더 축소)
\providecommand{\msepm}[2]{#1\,\ensuremath{\pm}\,{\fontsize{5.4pt}{6.1pt}\selectfont #2}}

% slightly smaller bold for Case entries (인덱스명은 그대로, 글자 크기만 축소)
\providecommand{\casebf}[1]{{\fontsize{6.8pt}{7.6pt}\selectfont\textbf{#1}}}

% header cell macro (multirow + vertical centering tweak)
\newlength{\TwoRowHeadRaise}
\setlength{\TwoRowHeadRaise}{%
  \dimexpr -0.5\aboverulesep - 0.5\belowrulesep - 0.5\lightrulewidth \relax}
\providecommand{\headcell}[1]{%
  \multirow{2}{*}[\TwoRowHeadRaise]{\centering\arraybackslash \textbf{#1}}%
}

% (글자/세로 간격만) 전체 폰트 및 행간 축소
{\fontsize{7.2pt}{8.0pt}\selectfont
\renewcommand{\arraystretch}{1.00} % 기존 1.13 → 1.00 (세로 간격 축소)
\setlength{\tabcolsep}{4.0pt}      % 좌우 간격은 그대로 유지

\resizebox{0.95\textwidth}{!}{%
\begin{tabular}{
  >{\centering\arraybackslash}m{0.18\textwidth}   % Case
  >{\centering\arraybackslash}m{0.115\textwidth}  % Preprocessing (X/O)
  >{\centering\arraybackslash}m{0.145\textwidth}  % Auxiliary network (X/O)
  >{\centering\arraybackslash}m{0.13\textwidth}   % #Parameters
  >{\centering\arraybackslash}m{0.12\textwidth}   % Epochs per case - Preprocessing
  >{\centering\arraybackslash}m{0.13\textwidth}   % Epochs per case - Training
  >{\centering\arraybackslash}m{0.18\textwidth}   % MSE ± σ
}
\specialrule{1.2pt}{0pt}{0pt} % outer top rule

% ---------- Header ----------
\headcell{Case} &
\headcell{\shortstack{Pre-\\processing}} &
\headcell{\shortstack{Auxiliary\\network}} &
\headcell{\#Parameters} &
\multicolumn{2}{c}{\textbf{Epochs per case}} &
\headcell{MSE $\,\pm\,$ $\sigma$} \\
\cmidrule(lr){5-6}
& & & & \textbf{Preprocessing} & \textbf{Training} & \\

\cmidrule[0.9pt](r){1-1}\cmidrule[0.9pt](l){2-7}

% ---------- Body ----------
\multirow{2}{*}{\casebf{Hyper-LR-PINN \cite{cho2023hypernetwork}}} &
\multirow{2}{*}{X} &
\multirow{2}{*}{O} &
\multirow{2}{*}{28{,}202} &
\multirow{2}{*}{0} &
20{,}000 & \msepm{4.27E-03}{0.0035} \\
& & & & & 10{,}000 & \msepm{1.54E-02}{0.0202} \\
\cmidrule[0.6pt](r){1-1}\cmidrule[0.6pt](l){2-7}

\multirow{2}{*}{\casebf{P$^2$INN \cite{cho2024ppinn}}} &
\multirow{2}{*}{X} &
\multirow{2}{*}{O} &
\multirow{2}{*}{126{,}338} &
\multirow{2}{*}{0} &
20{,}000 & \msepm{1.18E-02}{0.0145} \\
& & & & & 10{,}000 & \msepm{5.22E-02}{0.0468} \\
\cmidrule[0.6pt](r){1-1}\cmidrule[0.6pt](l){2-7}

\casebf{\shortstack{LAM\mbox{-}PINN\\(Ours)}} &
O & X & 30{,}021 & 100 & 200/100 & \msepm{1.14E-03}{0.0021} \\
\specialrule{1.2pt}{0pt}{0pt} % outer bottom rule
\end{tabular}
} % end resizebox
} % end fontsize group
\end{table*}
\endgroup
% =============================== End of Table 6 ============================================

\begin{figure}[t!]
\centering
\includegraphics[width=0.8\linewidth]{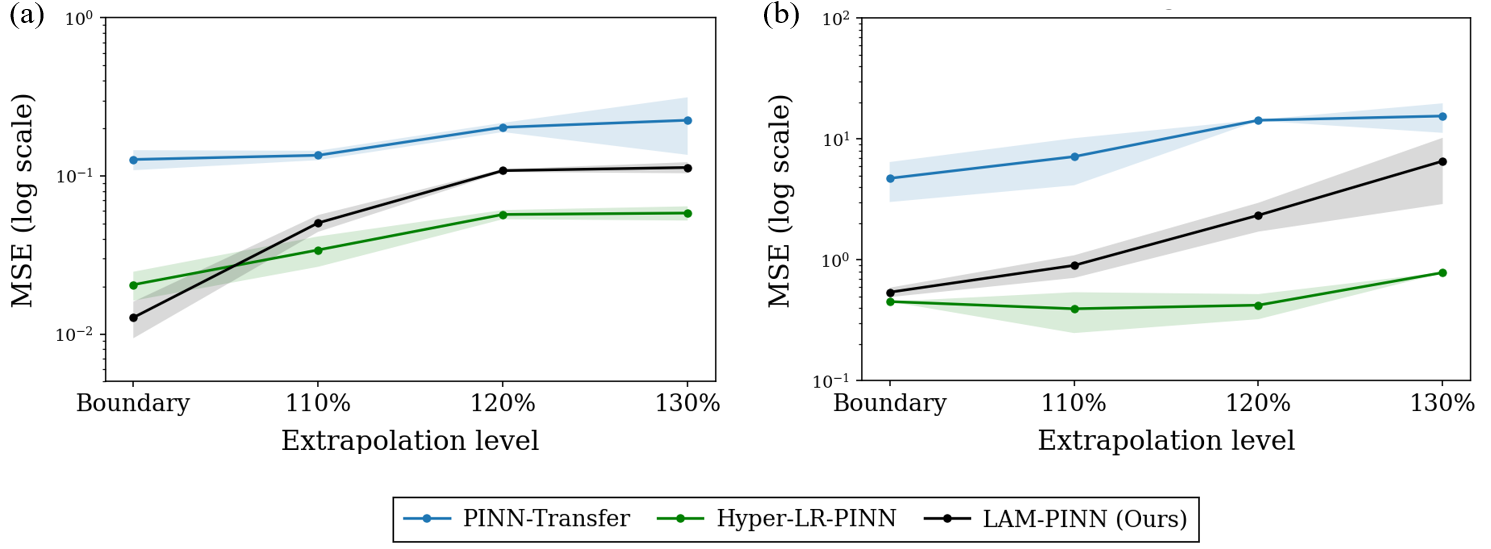}
% ▼ 이 구문을 통해 Figure 13: 번호와 텍스트 전체를 파란색으로 설정합니다.
% \captionsetup{labelfont={color=blue}, textfont={color=blue}}
\caption{
OOD extrapolation performance on (a) Helmholtz and (b) Burgers' benchmarks. Transfer MSE is evaluated as task parameters extend to 110\%--130\% of the DoE boundary. Curves show the average across five independent seeds; shaded bands indicate $\pm$ SD.
}
\label{fig:fig13}
\end{figure}

\subsubsection{Out-of-distribution extrapolation analysis}

To evaluate the robustness of LAM-PINN beyond the predefined task distribution, we conduct an out-of-distribution (OOD) extrapolation test on the Helmholtz and Burgers' benchmarks. OOD tasks are constructed by extending the design variables to 110\%, 120\%, and 130\% of the original DoE boundary, while keeping the transfer protocol identical to the main experiments. We compare against PINN-Transfer and Hyper-LR-PINN, representing \textit{i}) direct fine-tuning and \textit{ii}) auxiliary-network-based generalization, respectively; Hyper-LR-PINN was among the strongest baselines in our main experiments.

As shown in Fig.~\ref{fig:fig13}, LAM-PINN achieves low error near the DoE boundary and consistently outperforms PINN-Transfer. However, a distinct divergence is observed as the extrapolation magnitude grows toward 130\%. Specifically, Hyper-LR-PINN exhibits a milder MSE increase, whereas LAM-PINN degrades more noticeably. This behavior is consistent with our formulation, where LAM-PINN forms a finite set of cluster-specialized subnetworks from a small in-range task set and reuses them via routing. Consequently, under strong extrapolation, the learned modules may be insufficient to represent entirely new physical regimes.

% \begin{figure}[t!]
% \centering
% \includegraphics[width=0.8\linewidth]{Fig.13.PNG}
% \caption{
% OOD extrapolation performance on (a) Helmholtz and (b) Burgers' benchmarks. Transfer MSE is evaluated as task parameters extend to 110\%--130\% of the DoE boundary. Curves show the average across five independent seeds; shaded bands indicate $\pm$ SD.
% }
% \label{fig:fig13}
% \end{figure}

\section{Discussion}

Our results suggest that explicitly modeling task-specific learning dynamics can mitigate heterogeneity-induced negative transfer in PINNs when only a limited number of tasks is available. Compared to prior studies that vary only one or two parameters or mainly adjust PDE coefficients, we construct a three-factor, three-level full-factorial DoE by jointly varying three task-defining configurations, including PDE coefficients and ICs/BCs-related parameters. This setting is more susceptible to task heterogeneity and better reflects practical design-space exploration, where interpolation within a prescribed domain is often the primary objective. Within this in-range DoE setting, LAM-PINN yields low transfer errors on unseen tasks, including low-affinity cases where global initialization baselines are prone to negative transfer. This gain stems from task clustering via affinity signals from brief transfer sessions ($<5\%$ of typical PINN convergence) and selectively routing to specialized early-layer modules within a shared meta network.

The OOD extrapolation study (Fig.~\ref{fig:fig13}) also clarifies a limitation: as test conditions move beyond the DoE boundary, the error tends to increase, indicating reduced robustness under strong extrapolation. This is consistent with our formulation---LAM-PINN composes a finite set of modules learned from a small in-range task set, which may provide insufficient coverage when the target regime deviates substantially from the clustered task distribution. In contrast, auxiliary-network-based approaches, although typically requiring additional training and compute in our setting, may extrapolate more smoothly by learning a parameter-conditioned mapping that varies smoothly with the conditioning variables. This highlights a practical trade-off: LAM-PINN prioritizes efficient handling of in-range heterogeneity through task grouping, whereas more compute-intensive conditional generalization can be advantageous under larger distribution shifts.

\section{Conclusion}

This study introduced LAM-PINN, a compositional meta-learning framework that uses learning-affinity-based task representations and modular subnetworks to mitigate task heterogeneity in PINNs. By assigning shared physics to a global meta network and allocating task-specific variations to dedicated subnetworks, LAM-PINN reframes task heterogeneity as an opportunity for modular reuse rather than as a limitation. Under a $3{\times}3{\times}3$ DoE setting with three jointly varied task-defining configurations, LAM-PINN achieves improved transfer performance on unseen in-range tasks with a small preprocessing overhead. Across multiple PDE benchmarks, including higher-dimensional and geometrically irregular cases, LAM-PINN achieves substantially lower transfer error on unseen tasks than pre-trained, meta-learned, and auxiliary-network-based baselines. This gain is enabled by task representations derived from brief transfer sessions that incur only a modest preprocessing overhead, less than 5\% of a standard training budget.

Several limitations and future directions remain. First, the OOD extrapolation test (Fig.~\ref{fig:fig13}) shows that performance can degrade as tasks move farther beyond the DoE boundary, motivating extensions for improved robustness under strong extrapolation. Second, while adaptive $\lambda$-routing provides practical robustness to moderate misspecification of the number of clusters, probabilistic formulations (e.g., Bayesian mixture models) may further improve automated cluster selection and reliable routing initialization. Finally, scaling to highly complex industrial geometries may benefit from complementary mechanisms such as coordinate transformations or localized feature encodings, potentially combined with adaptive module expansion to better accommodate distribution shifts.

\medskip
\noindent $\textbf{Code Availability}$ The code and configuration files for the LAM-PINN reference implementation are publicly available at https://github.com/bc0322/LAM-PINN.

\medskip
\noindent $\textbf{Acknowledgement}$ This research was supported by the Institute of Information \& Communications Technology Planning \& Evaluation (IITP) grant, funded by the Korea government (MSIT) (No. RS-2019-II190079 (Artificial Intelligence Graduate School Program (Korea University)), and by the Korea Institute for Advancement of Technology (KIAT) grant, funded by the Ministry of Trade, Industry and Energy (MOTIE) (No. RS-2025-12572968).

\bibliographystyle{elsarticle-num}
\bibliography{reference}

% \clearpage
\appendix
\renewcommand{\thefigure}{A.\arabic{figure}}
\setcounter{figure}{0}
\setcounter{equation}{0}

\section{Experimental Setup for Each PDE}

\noindent Unless otherwise noted, method-specific optimization budgets, parameter counts, and learning-rate settings are summarized in Table~\ref{tab:baseline_configs}. The paragraphs below therefore focus on PDE formulations, DoE variables, selected LAM-PINN architectures, and collocation-point settings.

\medskip
\noindent\textbf{Helmholtz equation.} The 2D Helmholtz equation is defined by
\begin{equation}
u_{xx} + u_{yy} + \frac{u}{B^2} + \frac{u}{C^2} = 0
\end{equation}
with exact solution \(u(x,y)=A\sin(x/B)\sin(y/C)\) on \((x,y)\in[-30,30]^2\). The DoE variables are \(A\in[1,13]\), \(B\in[2,12]\), and \(C\in[3,11]\), and the reference pre-training task uses \(A=B=C=7\). For LAM-PINN, preprocessing uses a 50-epoch short transfer session on the 27 DoE tasks, with IN \([2,10,10]\), MN \([10,10,10,1]\), and \(K=3\) clusters (four INs including \(\theta_{\mathrm{IN}}^0\)). Optimization uses Adam with a 100/50 phase-wise budget, and 10{,}000 collocation points are sampled on a \(100\times100\) grid.

\medskip
\noindent\textbf{Burgers' equation.} We use the 1D Burgers' equation
\begin{equation}
\frac{\partial u}{\partial t} + \alpha \frac{\partial}{\partial x}(u^2) = \nu \frac{\partial^2 u}{\partial x^2},
\end{equation}
on \(x\in[-1,1]\) and \(t\in[0,1]\), with initial condition \(u(x,0)=-A\sin(\pi x)\). The DoE variables are \(\alpha\in[0.1,2]\), \(\nu\in[0.005,0.5]\), and \(A\in[0.5,10]\), and the reference pre-training task uses \((\alpha,\nu,A)=(1,0.03,5)\). For LAM-PINN, preprocessing uses 500 transfer epochs on the 27 DoE tasks, with IN \([2,20,20,20,20]\), MN \([20,20,20,20,20,1]\), and \(K=5\) clusters (six INs including \(\theta_{\mathrm{IN}}^0\)). We employ L-BFGS optimization with a 1000/100 phase-wise budget, 50{,}000 boundary and 100{,}000 interior collocation points, and finite-difference/Runge--Kutta reference solutions for error evaluation.

% =============================== Table A.1: Start ===============================
\begin{table*}[t]
\singlespacing
\centering
% Requires: \usepackage{booktabs, array, multirow, caption}
% \captionsetup{font=small, labelfont={color=blue}, textfont={color=blue}}
\setcounter{table}{0}
\renewcommand{\thetable}{A.\arabic{table}}

\caption{Method-specific settings and computational budgets across compared methods.}
\label{tab:baseline_configs}
% \color{blue}

\newcolumntype{C}[1]{>{\centering\arraybackslash}m{#1}}
\newcolumntype{L}[1]{>{\raggedright\arraybackslash}m{#1}}

{\fontsize{6.6pt}{7.8pt}\selectfont
\renewcommand{\arraystretch}{1.14}
\setlength{\tabcolsep}{2.6pt}

\resizebox{\textwidth}{!}{%
\begin{tabular}{
  C{2.25cm}
  C{1.00cm}
  C{1.10cm} C{1.28cm} C{1.24cm}
  C{1.00cm} C{1.00cm} C{1.20cm}
  C{4.05cm}
  C{1.15cm} C{1.15cm}
  C{1.85cm}
}
\specialrule{1.2pt}{0pt}{1.2pt}

\multirow{2}{*}{\parbox[c]{2.25cm}{\centering\textbf{Method}}} &
\multirow{2}{*}{\parbox[c]{1.00cm}{\centering\textbf{Preproc.}\\\textbf{stage}}} &
\multicolumn{3}{c}{\parbox[c]{3.62cm}{\centering\textbf{Optimization budget}\\\textbf{[epochs/task]}}} &
\multicolumn{3}{c}{\parbox[c]{3.20cm}{\centering\textbf{\#Parameters}}} &
\multirow{2}{*}{\parbox[c]{4.05cm}{\centering\textbf{Method-specific setting}}} &
\multicolumn{2}{c}{\parbox[c]{2.30cm}{\centering\textbf{Learning rate (init.)}}} &
\multirow{2}{*}{\parbox[c]{1.85cm}{\centering\textbf{LR schedule}}} \\
\cmidrule(lr){3-5}\cmidrule(lr){6-8}\cmidrule(lr){10-11}

& &
\textbf{Helmholtz} & \textbf{Burgers'} & \shortstack{\textbf{Linear}\\\textbf{elasticity}} &
\textbf{Helmholtz} & \textbf{Burgers'} & \shortstack{\textbf{Linear}\\\textbf{elasticity}} &
& \textbf{Train} & \textbf{Transfer} & \\
\specialrule{0.8pt}{0pt}{0pt}

\textbf{PINN \cite{raissi2019physics}}
  & -- & 5,000 & 30,000 & 5,000
  & 371 & 3,021 & 16,962
  & -- & 2.00E-03 & 2.00E-03 & Reduce-on-Plateau \\

\textbf{MAML \cite{finn2017model}}
  & -- & 150 & 1,100 & 300
  & 371 & 3,021 & 16,962
  & Inner/outer loop training & 2.00E-03 & 2.00E-03 & Reduce-on-Plateau \\

\textbf{ConML \cite{wu2025learning}}
  & -- & 150 & 1,100 & 300
  & 371 & 3,021 & 16,962
  & MAML + task-level contrastive learning & 2.00E-03 & 2.00E-03 & Reduce-on-Plateau \\

\textbf{MAD \cite{huang2022meta}}
  & -- & 150 & 1,100 & 300
  & 675 & 3,245 & 17,986
  & per-task latent code $z$ + shared decoder & 1.00E-03 & 2.00E-03 & Step decay \\

\textbf{DATS-w \cite{toloubidokhti2024dats}}
  & -- & 150 & 1,100 & 300
  & 1,225 & 3,773 & 18,418
  & MAD + adaptive task-loss weighting & 2.00E-03 & 2.00E-03 & Reduce-on-Plateau \\

\textbf{Hyper-LR-PINN \cite{cho2023hypernetwork}}
  & -- & 20,000 & 20,000 & 20,000
  & 28,202 & 28,202 & 28,202
  & Hypernet: 3-layer + 3 layer-wise head & 1.00E-03 & 2.50E-04 & -- \\

\textbf{P$^2$INN \cite{cho2024ppinn}}
  & -- & 20,000 & 20,000 & 20,000
  & 76,851 & 76,851 & 126,338
  & EqEncoder + CoordEncoder + Residual Decoder & 1.00E-03 & 1.00E-03 & -- \\

\textbf{LAM-PINN (ours)}
  & $\checkmark$ & 50* / 150 & 500* / 1,100 & 100* / 300
  & 794 & 9,626 & 30,021
  & Learning-affinity clustering + modular subnetworks ($\lambda$-routing)
  & 2.00E-03 & 2.00E-03 & Reduce-on-Plateau \\

\specialrule{1.2pt}{0pt}{0pt}
\end{tabular}%
}% end resizebox
}% end fontsize group
\par\vspace{0pt}
\noindent\hfill{\fontsize{5.2pt}{6.0pt}\selectfont\textit{Note:} * denotes the preprocessing budget.}
\vspace{0pt}
\end{table*}
% =============================== Table A.1: End ===============================

\medskip
\noindent\textbf{Linear elasticity equation and irregular geometry}

\begin{figure}[t!]
\centering
\includegraphics[width=0.7\linewidth]{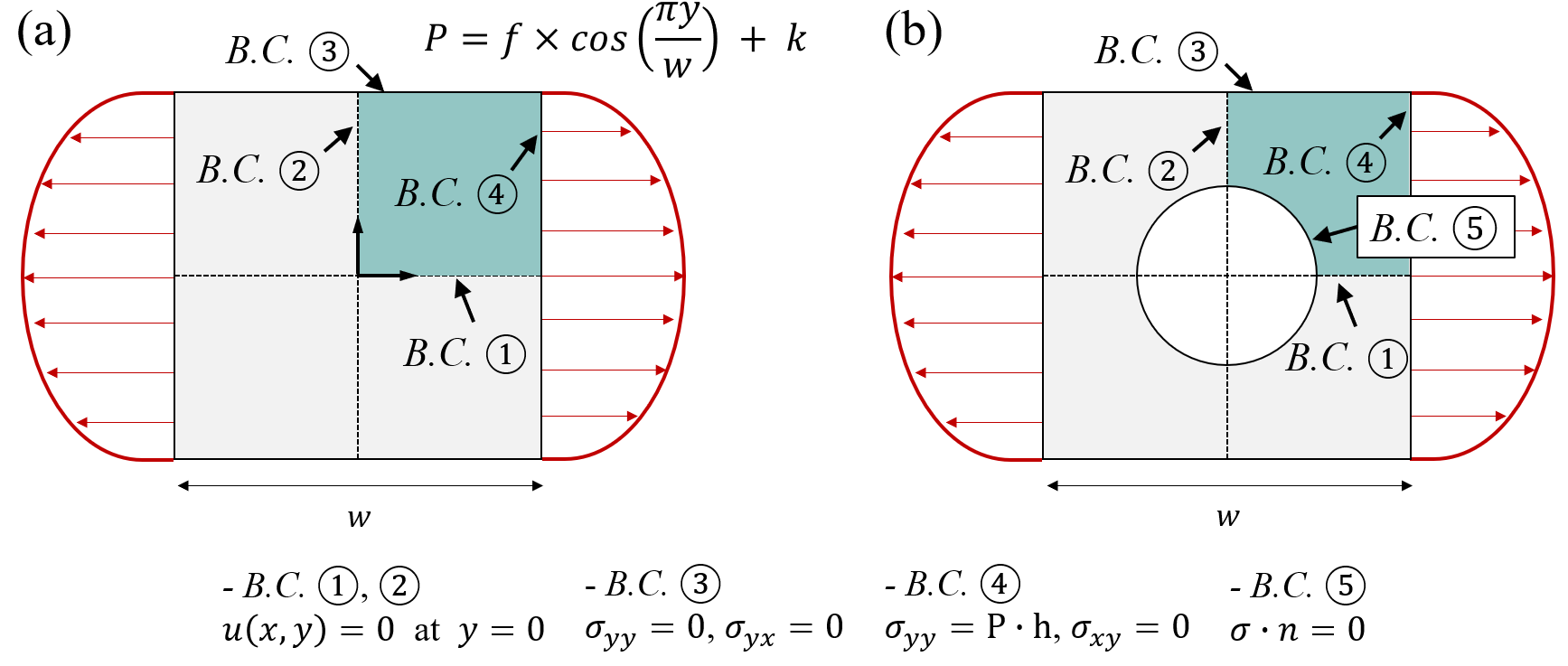}
\caption{
Schematic of the applied load and BCs on the solid: (a) standard 1/4 plate and (b) irregular geometry with a circular hole. A 1/4 symmetry model was used to analyze the behavior across the entire domain of the target solid.
}
\label{fig:fig.A.1}
% \vspace{-0.8em}
\end{figure}

\noindent\textbf{Common physical model and settings.}
The governing plane-stress linear elasticity equations are
\begin{equation}
\frac{E}{1 - \nu^2} \left( \frac{\partial^2 u}{\partial x^2} + \frac{1 - \nu}{2} \frac{\partial^2 u}{\partial y^2} + \frac{1 + \nu}{2} \frac{\partial^2 v}{\partial x \partial y} \right) = 0,
\end{equation}
\begin{equation}
\frac{E}{1 - \nu^2} \left( \frac{\partial^2 v}{\partial y^2} + \frac{1 - \nu}{2} \frac{\partial^2 v}{\partial x^2} + \frac{1 + \nu}{2} \frac{\partial^2 u}{\partial x \partial y} \right) = 0.
\end{equation}
Here, \(u(x,y)\) and \(v(x,y)\) denote the displacements in the \(x\)- and \(y\)-directions, respectively; \(E\) is Young's modulus, and \(\nu\) is Poisson's ratio. A \(1/4\) symmetry model and boundary conditions, as shown in Fig.~\ref{fig:fig.A.1}, were used to simulate the full domain of the target solid. A load \(P = f \cos\!\left(\frac{\pi y}{w}\right) + k\) was applied to the right boundary. The DoE variables were \(E \in [50,150]\), \(f \in [0.1,2]\), and \(k \in [0,10]\).  Verification of predictions was conducted using the commercial finite element method program \textsc{ABAQUS} 2023. Figure~\ref{fig:fig.A.1} illustrates the applied boundary conditions and loading.

\medskip
\noindent\textbf{Standard plate.}
The main experiments use a quarter-symmetric rectangular plate without holes. The reference pre-training task is \((E,f,k)=(100,1,5)\). For LAM-PINN, preprocessing uses 100 transfer epochs on the 27 DoE tasks, with IN \([2,64,64]\), MN \([64,64,64,64,2]\), and \(K=3\) clusters (four INs including \(\theta_{\mathrm{IN}}^0\)). Optimization uses Adam with a 200/100 phase-wise budget. We use 5{,}000 boundary and 10{,}000 interior collocation points.

\medskip
\noindent\textbf{Plate with a circular hole.}
To evaluate irregular geometry and stress concentration, we additionally consider a plate with a central circular hole of radius \(r=0.25w\). The constitutive model and loading are unchanged. For this setting, LAM-PINN is trained with a 2000/100 phase-wise budget, and 13{,}824 boundary plus 110{,}592 interior collocation points are used. PINN-Transfer and Hyper-LR-PINN are trained for 20{,}000 epochs in this geometry to ensure convergence.

\medskip
\noindent\textbf{3D Helmholtz Equation.} The governing 3D Helmholtz equation is defined as $\Delta u + u = f$ on the domain $\Omega = [-1, 1]^3$, with the exact solution given by $u(x, y, z) = \sin(a_1 \pi x)\sin(a_2 \pi y)\sin(a_3 \pi z)$.
The 3-factor, 3-level DoE for the 3D Helmholtz equation was constructed using parameters $a_1 \in [0.5, 1.0, 1.5]$, $a_2 \in [0.2, 1.1, 2.0]$, and $a_3 \in [1.0, 2.0, 3.0]$.

For the baseline models, a conventional PINN was trained for 10,000 epochs, while Hyper-LR-PINN was trained for 20,000 epochs. In the preprocessing phase, a brief transfer-learning session of 400 epochs was performed for each of the 27 DoE tasks to extract learning-affinity metrics. The LAM-PINN architecture used an Input Network (IN) of size $[3, 96]$ and a Meta Network (MN) of size $[96, 96, 96, 1]$. Model training consisted of Phase 1 (200 epochs) and Phase 2 (100 epochs), using the Adam optimizer with a learning rate of $10^{-3}$. A total of 110{,}592 interior and 13{,}824 boundary collocation points were used for training.

% =============================== Appendix B: End ===============================

% ===============================Table B.1: Start=====================================
\begin{table*}[t]
\singlespacing
\centering
\setcounter{table}{0}
\renewcommand{\thetable}{B.\arabic{table}}
\caption{Cluster validity and stability across 3 PDEs. For each PDE, we report the average silhouette score ($\pm$ SD) and the average pairwise ARI over 20 seeds.}
\label{tab:clustering_stats}

% Column type definition
\newcolumntype{M}[1]{>{\centering\arraybackslash}m{#1}}

% mean ± sd (keep as-is)
\newcommand{\meansd}[2]{\ensuremath{#1\pm{\fontsize{6pt}{7.2pt}\selectfont #2}}}
\newcommand{\meansdB}[2]{\ensuremath{\mathbf{#1}\pm{\fontsize{6pt}{7.2pt}\selectfont \mathbf{#2}}}}

% ↓ Font one step smaller + tighter vertical spacing
{\fontsize{6pt}{7.2pt}\selectfont
\renewcommand{\arraystretch}{1.00}  % 1.15 -> 1.00 (tighter rows)
\setlength{\tabcolsep}{5pt}        % keep horizontal spacing as-is

% tighten booktabs vertical gaps (rule spacing)
\setlength{\aboverulesep}{0.3pt}
\setlength{\belowrulesep}{0.3pt}

\resizebox{0.95\textwidth}{!}{%
\begin{tabular}{
    M{0.09\textwidth}
    M{0.14\textwidth} M{0.08\textwidth}
    M{0.14\textwidth} M{0.08\textwidth}
    M{0.14\textwidth} M{0.08\textwidth}
}
% reduce the extra below-space of the top rule (was 1.2pt)
\specialrule{1.2pt}{0pt}{0.7pt}
\multirow{2}{*}[-0.2ex]{\shortstack[c]{\fontsize{5pt}{5.8pt}\selectfont\bfseries Number of\\ Cluster $K$}} &

% \multirow{2}{*}[-0.2ex]{\shortstack{\textbf{Number}\\\textbf{of}\\\textbf{Cluster $K$}}} &
\multicolumn{2}{c}{\textbf{Helmholtz}} &
\multicolumn{2}{c}{\textbf{Burgers'}} &
\multicolumn{2}{c}{\textbf{Linear elasticity}} \\
\cmidrule(lr){2-3}\cmidrule(lr){4-5}\cmidrule(lr){6-7}
& \textbf{Silhouette} & \textbf{ARI}
& \textbf{Silhouette} & \textbf{ARI}
& \textbf{Silhouette} & \textbf{ARI} \\
\cmidrule(lr){1-1}\cmidrule(lr){2-2}\cmidrule(lr){3-3}
\cmidrule(lr){4-4}\cmidrule(lr){5-5}\cmidrule(lr){6-6}\cmidrule(lr){7-7}

2 & \meansd{0.364}{0.004} & 0.98
  & \meansd{0.194}{0.023} & 0.29
  & \meansd{0.288}{0.027} & 0.97 \\
3 & \meansdB{0.312}{0.002} & \textbf{1.00}
  & \meansd{0.226}{0.016} & 0.54
  & \meansdB{0.266}{0.022} & \textbf{1.00} \\
4 & \meansd{0.291}{0.022} & 0.96
  & \meansd{0.244}{0.018} & 0.70
  & \meansd{0.251}{0.018} & 0.73 \\
5 & \meansd{0.309}{0.030} & 0.86
  & \meansdB{0.263}{0.023} & \textbf{0.81}
  & \meansd{0.241}{0.023} & 0.76 \\
6 & \meansd{0.297}{0.021} & 0.78
  & \meansd{0.262}{0.021} & 0.80
  & \meansd{0.244}{0.025} & 0.73 \\

\specialrule{1.2pt}{0pt}{0pt}
\end{tabular}
} % end resizebox
} % end fontsize group
\end{table*}
% ===============================Table B.1: End======================================

% =========================== Final Transfer MSE (1000 epochs) — small font, 5-dec std ===========

\setcounter{table}{1}
\renewcommand{\thetable}{B.\arabic{table}}
\begin{table*}[t]
\singlespacing
\centering
% Requires: \usepackage{booktabs, array, caption, setspace, xcolor}
% \captionsetup{labelfont={bf,color=blue}, textfont={color=blue}}

\caption{Final transfer performance of LAM-PINN on 10 unseen 3D Helmholtz tasks. The transfer session indicates the fraction of the full training budget used to form clusters, and the table reports the average MSE $\pm$ SD after 1000 epochs.}
\label{tab:final_transfer_mse}

% Centered column type (horizontal & vertical centering)
\newcolumntype{M}[1]{>{\centering\arraybackslash}m{#1}}
% mean ± sd macro (std in slightly smaller font)
\newcommand{\meansd}[2]{\ensuremath{#1\pm{\fontsize{6pt}{7.5pt}\selectfont #2}}}

{\fontsize{7pt}{8.5pt}\selectfont
\renewcommand{\arraystretch}{1.25}
\setlength{\tabcolsep}{6pt}

\resizebox{0.85\textwidth}{!}{%
\begin{tabular}{
  M{0.26\textwidth}  % Transfer session
  M{0.59\textwidth}  % Final MSE ± σ
}
\specialrule{1.2pt}{0pt}{1.0pt}

\textbf{Transfer session} & \textbf{MSE $\pm$ $\sigma$ (1000 epochs)} \\
\cmidrule(lr){1-1}\cmidrule(lr){2-2}

\textbf{2\%}  & \meansd{0.00395}{0.00065} \\
\textbf{4\%}  & \meansd{0.00403}{0.00090} \\
\textbf{10\%} & \meansd{0.00423}{0.00086} \\

\specialrule{1.2pt}{0pt}{0pt}
\end{tabular}
} % end resizebox
} % end fontsize group
% \vspace{1cm}
\end{table*}
% ======================== End: Final Transfer MSE (1000 epochs) — small font, 5-dec std =====

% =============================== Appendix B: Start ===============================
% (Place this block AFTER Appendix A.3 in your existing code.)

% Reset numbering to Appendix B
\renewcommand{\thefigure}{B.\arabic{figure}}
\renewcommand{\thetable}{B.\arabic{table}}
\setcounter{figure}{0}
\setcounter{table}{0}

\section*{Appendix B. Clustering validation protocol and selection rule}
\label{app:cluster_validation}

\noindent\textbf{Task representation.}
We cluster the task embeddings \(f_a\) (Eq.~(4)), formed by concatenating PDE parameters with three loss-derived metrics from a short transfer session. Features are transformed by \(\log(1+Z)\) and standardized (Z-score) before applying \(k\)-means.

\medskip
\noindent\textbf{Evaluation protocol.}
For each PDE and \(K \in \{2,\ldots,6\}\), \(k\)-means is run 20 times with different random seeds. We report \textit{i}) silhouette scores (mean \(\pm\) SD) and \textit{ii}) seed-level stability via the average pairwise ARI. Numerical summaries for the representative short-session setting are given in Table~\ref{tab:clustering_stats}.

\begin{figure}[t!]
\centering
\includegraphics[width=0.8\linewidth]{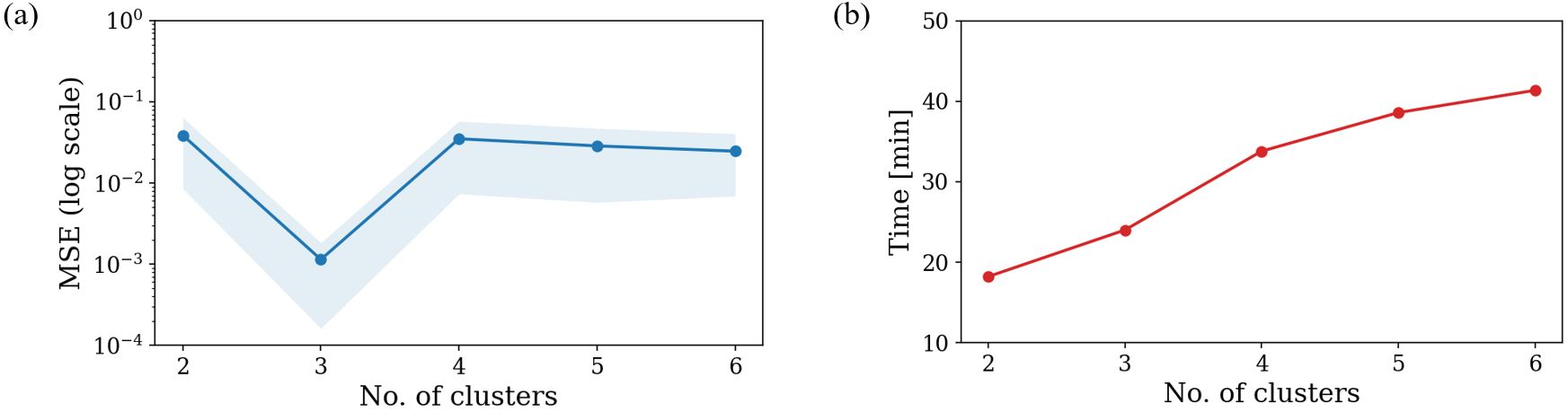}
% \captionsetup{labelfont={color=blue}, textfont={color=blue}}
\caption{
Sensitivity to the number of clusters \(K\) on the linear-elasticity benchmark under the same DoE setting:  (a) average transfer MSE on 10 unseen tasks and (b) training time versus  \(K\).
}

\label{fig:fig.B.1}
% \vspace{-0.8em}
\end{figure}

\medskip
\noindent\textbf{Selection rule for \(K\).}
\hl{Because optimizing a single internal validity index in isolation can bias the choice toward smaller \(K\) on some datasets, we select \(K\) by jointly considering cluster separation (silhouette) and seed-level stability (ARI), and then fix one \(K\) per PDE for all downstream experiments (Helmholtz/Elasticity: \(K{=}3\); Burgers: \(K{=}5\)). Fig.~\ref{fig:fig.B.1} complements the internal clustering indices with a representative downstream sensitivity analysis on the linear-elasticity benchmark. It shows that larger \(K\) does not guarantee better transfer performance, while increasing training cost, supporting \(K{=}3\) as a practical trade-off.}

\medskip
\noindent\textbf{Sensitivity to transfer-session budget and cluster granularity.}
We also analyze how the transfer-session budget used to construct the learning-affinity metrics influences both clustering and downstream performance. Using the 3D Helmholtz equation, we form clusters based on transfer sessions corresponding to 2\%, 4\%, and 10\% of the full-convergence iterations, train LAM-PINN with these clusterings, and evaluate transfer to ten unseen tasks (Table~\ref{tab:final_transfer_mse}). The three settings yield closely matched final errors, indicating that LAM-PINN is practically insensitive to moderate variations in the transfer-session budget and, by extension, to mild over- or under-estimation of the number of clusters. Together with the label-disagreement analysis in Table~\ref{tab:helmholtz_combined}, these results support the use of a short transfer budget of less than 5\% as a stable default in resource-constrained environments.

\renewcommand{\thetable}{C.\arabic{table}}
\setcounter{table}{0}
\renewcommand{\theequation}{C.\arabic{equation}}
\setcounter{equation}{0}

\begin{revblue}
\section*{Appendix C. Fixed-benchmark seed sensitivity and uncertainty}
\label{app:seed_sensitivity}

We performed an additional seed-sensitivity check for the fixed 10-task benchmark setting. Representative benchmark evaluations were repeated over 10 independent runs with different random seeds while keeping the unseen-task sets unchanged. Table~\ref{tab:seed_sensitivity} reports the resulting mean MSE \(\pm\) SD across seeds for representative methods on the Helmholtz and linear-elasticity benchmarks, showing that the main comparative trend is not driven by a particular fixed seed.

We additionally quantified uncertainty in the main fixed-benchmark comparative gains. Following the same experimental setup, the MSE for each unseen task was averaged over the 10 independent runs. With the resulting 10 paired task-level mean MSE values, the relative reduction of LAM-PINN over a baseline was defined as
\begin{equation}
\mathrm{Reduction}(\%) = 100 \times\left(1-\frac{\overline{\mathrm{MSE}}_{\text{ours}}}{\overline{\mathrm{MSE}}_{\text{base}}}\right),
\end{equation}
where \(\overline{\mathrm{MSE}}\) denotes the task-level MSE averaged over runs. We then estimated 95\% confidence intervals by paired bootstrap resampling (10{,}000 resamples) and computed \(p\)-values using two-sided paired Wilcoxon signed-rank tests. Table~\ref{tab:ci_main_benchmarks} reports, for each benchmark, three representative baselines from the most competitive comparisons.
\end{revblue}

% =============================== Table C.1: Start ===============================
\begin{table*}[t]
\singlespacing
\centering
% \captionsetup{labelfont={color=blue}, textfont={color=blue}}

\caption{Seed sensitivity on fixed 10-task Helmholtz and linear-elasticity benchmarks. Entries report mean MSE $\pm$ SD over 10 independent run seeds. Group~A denotes the top-5 tasks with the highest learning-affinity metrics, while Group~B denotes the remaining bottom-5 tasks.}
\label{tab:seed_sensitivity}
% \color{blue}

{\fontsize{7.6pt}{9.0pt}\selectfont
\renewcommand{\arraystretch}{1.22}
\setlength{\tabcolsep}{4.5pt}

\newcommand{\seedsubm}[1]{{\fontsize{5.6pt}{6.4pt}\selectfont #1}}
\newcommand{\seedsubs}[1]{{\fontsize{5.1pt}{5.9pt}\selectfont #1}}
\newcommand{\sdmant}[1]{{\fontsize{7.2pt}{8.2pt}\selectfont #1}}
\newcommand{\expfmt}[2]{#1\text{\seedsubm{#2}}}
\newcommand{\expfmtsd}[2]{\text{\sdmant{#1}}\text{\seedsubs{#2}}}
\newcommand{\seedmeansd}[4]{\ensuremath{\expfmt{#1}{#2}\pm\expfmtsd{#3}{#4}}}

\resizebox{\textwidth}{!}{%
\begin{tabular}{c c c c @{\hspace{10pt}} c c c}
\specialrule{1.2pt}{0pt}{1.2pt}

\multirow{3}{*}{\textbf{Method}} &
\multicolumn{6}{c}{\textbf{Average MSE} (\(\pm\sigma\))} \\
\cmidrule(lr){2-7}

& \multicolumn{3}{c}{\textbf{Helmholtz}} &
  \multicolumn{3}{c}{\textbf{Linear Elasticity}} \\
\cmidrule(lr){2-4}\cmidrule(lr){5-7}

& \textbf{10 Tasks} & \textbf{A Group} & \textbf{B Group}
& \textbf{10 Tasks} & \textbf{A Group} & \textbf{B Group} \\
\cmidrule(l){1-1}\cmidrule(l){2-2}\cmidrule(l){3-3}\cmidrule(l){4-4}
\cmidrule(l){5-5}\cmidrule(l){6-6}\cmidrule(l){7-7}

\textbf{ConML \cite{wu2025learning}} &
\seedmeansd{1.48}{E+00}{1.52}{E-01} &
\seedmeansd{2.93}{E+00}{3.03}{E-01} &
\seedmeansd{3.58}{E-02}{3.97}{E-03} &
\seedmeansd{2.93}{E-02}{1.57}{E-03} &
\seedmeansd{5.47}{E-02}{3.45}{E-03} &
\seedmeansd{3.86}{E-03}{2.98}{E-04} \\

\textbf{DATS-w \cite{toloubidokhti2024dats}} &
\seedmeansd{2.63}{E+00}{7.87}{E-02} &
\seedmeansd{4.90}{E+00}{1.13}{E-01} &
\seedmeansd{3.66}{E-01}{4.79}{E-02} &
\seedmeansd{3.90}{E-03}{1.35}{E-04} &
\seedmeansd{6.06}{E-03}{2.74}{E-04} &
\seedmeansd{1.74}{E-03}{3.67}{E-05} \\

\textbf{Hyper-LR-PINN \cite{cho2023hypernetwork}} &
\seedmeansd{1.85}{E+00}{1.64}{E-02} &
\seedmeansd{3.58}{E+00}{3.18}{E-02} &
\seedmeansd{1.22}{E-01}{5.29}{E-05} &
\seedmeansd{4.28}{E-03}{3.24}{E-04} &
\seedmeansd{6.74}{E-03}{4.90}{E-04} &
\seedmeansd{1.82}{E-03}{3.58}{E-04} \\

\textbf{LAM-PINN (ours)} &
\seedmeansd{1.47}{E-01}{5.63}{E-03} &
\seedmeansd{2.71}{E-01}{1.49}{E-03} &
\seedmeansd{2.38}{E-02}{1.09}{E-02} &
\seedmeansd{1.08}{E-03}{7.54}{E-05} &
\seedmeansd{1.45}{E-03}{1.40}{E-04} &
\seedmeansd{7.03}{E-04}{8.92}{E-05} \\

\specialrule{1.2pt}{0pt}{0pt}
\end{tabular}
} % end resizebox
} % end fontsize
\end{table*}
% =============================== Table C.1: End ===============================

% =============================== Table C.2: Start ===========================================
\begingroup
\begin{table*}[t]
\centering
% \captionsetup{labelfont={color=blue}, textfont={color=blue}}
\caption{Uncertainty analysis for the main 10-task benchmark comparisons. Entries report reduction (\%), 95\% confidence intervals, and \(p\)-values for representative competitive baselines.}

\label{tab:ci_main_benchmarks}
% \color{blue}

\setlength{\aboverulesep}{0.35ex}
\setlength{\belowrulesep}{0.35ex}
\setlength{\cmidrulesep}{0.20ex}

{\fontsize{7.1pt}{7.9pt}\selectfont
\renewcommand{\arraystretch}{1.02}
\setlength{\tabcolsep}{4.2pt}

\resizebox{0.92\textwidth}{!}{%
\begin{tabular}{
  >{\centering\arraybackslash}m{0.16\textwidth}
  >{\centering\arraybackslash}m{0.24\textwidth}
  >{\centering\arraybackslash}m{0.13\textwidth}
  >{\centering\arraybackslash}m{0.23\textwidth}
  >{\centering\arraybackslash}m{0.12\textwidth}
}
\specialrule{1.2pt}{0pt}{0pt}

\textbf{PDE benchmark} &
\textbf{Baseline} &
\textbf{Reduction (\%)} &
\textbf{95\% CI} &
\textbf{\(p\)-value} \\
\cmidrule(lr){1-5}

\multirow{3}{*}{\textbf{Helmholtz}}
& ConML \cite{wu2025learning}
& 91.5
& [51.55, 93.19]
& 0.0273 \\
& MAD \cite{huang2022meta}
& 93.0
& [44.16, 94.83]
& 0.0488 \\
& Hyper-LR-PINN \cite{cho2023hypernetwork}
& 93.2
& [61.35, 95.82]
& 0.0371 \\
\cmidrule(lr){1-5}

\multirow{3}{*}{\textbf{Burgers'}}
& Hyper-LR-PINN \cite{cho2023hypernetwork}
& 80.4
& [41.35, 96.56]
& 0.0020 \\
& MAD \cite{huang2022meta}
& 85.2
& [22.81, 87.55]
& 0.0020 \\
& ConML \cite{wu2025learning}
& 88.6
& [67.76, 98.87]
& 0.0195 \\
\cmidrule(lr){1-5}

\multirow{3}{*}{\textbf{Linear elasticity}}
& Hyper-LR-PINN \cite{cho2023hypernetwork}
& 78.3
& [42.64, 91.47]
& 0.0273 \\
& DATS-w \cite{toloubidokhti2024dats}
& 82.1
& [57.46, 92.93]
& 0.0137 \\
& MAD \cite{huang2022meta}
& 93.7
& [90.75, 94.94]
& 0.0059 \\

\specialrule{1.2pt}{0pt}{0pt}
\end{tabular}
} % end resizebox
} % end fontsize
\end{table*}
\endgroup
% =============================== Table C.2: End ===========================================

%%       \cite{<label>} ==> [1]

\end{document}